\documentclass[journal]{IEEEtran}

\ifCLASSINFOpdf

\else

\fi

\usepackage{float}
\usepackage{amsmath,amssymb,amsfonts}
\usepackage{algorithmic}
\usepackage{graphicx}
\usepackage{textcomp}
\usepackage{float}
\usepackage{cite,url}
\usepackage{amsmath}
\usepackage{epstopdf} 

\usepackage{cleveref}
\usepackage{subfigure}
\usepackage[usestackEOL]{stackengine}

\begin{document}
	
	\newcommand{\subscript}[1]{$_{\text{#1}}$}
	
	\title{A Phase Variable Approach for Improved Rhythmic and Non-Rhythmic Control of a Powered Knee-Ankle Prosthesis}
	
	\author{Siavash Rezazadeh$^1$, David Quintero$^{2}$, Nikhil Divekar$^1$, Emma Reznick$^1$, Leslie Gray$^3$, and Robert D. Gregg$^{4}$
		\thanks{$^1$Department of Bioengineering, University of Texas at Dallas, Richardson, TX 75080, USA. $^2$School of Engineering, San Francisco State University, San Francisco, CA 94132, USA. $^3$Department of Health Care Sciences, University of Texas Southwestern Medical Center, Dallas, TX 75390, USA. $^4$Department of Electrical Engineering and Computer Science and the Robotics Institute, University of Michigan, Ann Arbor, MI 48109 USA. Contact emails: \tt\small siavash.rezazadeh@utdallas.edu, rdgregg@umich.edu}
		\thanks{This work was supported by the National Institute of Child Health \& Human Development of the NIH under Award Numbers DP2HD080349 and R01HD09477. This work was also supported by NSF Award 1734600. The content is solely the responsibility of the authors and does not necessarily represent the official views of the NIH or the NSF. R. D. Gregg holds a Career Award at the Scientific Interface from the Burroughs Wellcome Fund.
			
			$\copyright$ 2019 IEEE.  Personal use of this material is permitted.  Permission from IEEE must be obtained for all other uses, in any current or future media, including reprinting/republishing this material for advertising or promotional purposes, creating new collective works, for resale or redistribution to servers or lists, or reuse of any copyrighted component of this work in other works.} 
	}

	\maketitle
	
	\begin{abstract}
		Although there has been recent progress in control of multi-joint prosthetic legs for rhythmic tasks such as walking, control of these systems for non-rhythmic motions and general real-world maneuvers is still an open problem. In this article, we develop a new controller that is capable of both rhythmic (constant-speed) walking, transitions between speeds and/or tasks, and some common volitional leg motions. We introduce a new piecewise holonomic phase variable, which, through a finite state machine, forms the basis of our controller. The phase variable is constructed by measuring the thigh angle, and the transitions in the finite state machine are formulated through sensing foot contact along with attributes of a nominal reference gait trajectory. The controller was implemented on a powered knee-ankle prosthesis and tested with a transfemoral amputee subject, who successfully performed a wide range of rhythmic and non-rhythmic tasks, including slow and fast walking, quick start and stop, backward walking, walking over obstacles, and kicking a soccer ball. Use of the powered leg resulted in clinically significant reductions in amputee compensations for rhythmic tasks (including vaulting and hip circumduction) when compared to use of the take-home passive leg. In addition, considerable improvements were also observed in the performance for non-rhythmic tasks. The proposed approach is expected to provide a better understanding of rhythmic and non-rhythmic motions in a unified framework, which in turn can lead to more reliable control of multi-joint prostheses for a wider range of real-world tasks.
	\end{abstract}


	\IEEEpeerreviewmaketitle

\section{Introduction}
\IEEEPARstart{T}{he} vast majority of lower-limb amputees use mechanically passive prosthetic legs, which can only dissipate energy during locomotion. This limits the ability of amputees to efficiently perform various ambulation modes, particularly walking at variable speeds or slopes. Furthermore, the biomechanical compensations required to walk with these passive devices generally cause joint discomfort and back pain during daily usage \cite{Rabuffetti2005}. Powered prosthetic legs that provide actuation at the joints have the potential to improve amputee gait and eliminate these problems \cite{au2008powered,bellman2008sparky,ficanha2016cable,sup2008,Quintero2018}. However, they require sophisticated control strategies, especially for multi-joint legs, in order to perform various activities in a natural and safe manner \cite{Tucker2015}. 

From a biomechanics perspective, the human gait cycle can be divided into different phases (namely, stance and swing phase) and sub-phases (for example, weight acceptance, push-off, early swing, etc.), each serving a specific purpose in locomotion \cite{Winter2005}. This perspective was preserved in control design for powered lower-limb prostheses, which involves first detecting the correct sub-phase and then controlling that particular behavior of the prosthetic joints \cite{sup2008,Lenzi2014,simon2014,Lawson2014,Lawson2015}. The tuning has to be performed separately for each individual based on various physical parameters, for example, body mass, as well as functional parameters, for example, gait pattern. Due to the large number of parameters that need to be manually tuned, the process is typically arduous, often taking multiple hours for each subject \cite{simon2014}. Methods have been developed to automate this tuning process for a single joint (e.g., a knee prosthesis in \cite{Huang2019} and an ankle exoskeleton in \cite{Collins2017Science}), but further investigations are required to extend these methods to multi-joint prostheses.

To address these issues, recent approaches have parameterized the gait cycle over a phase variable, i.e., a monotonic signal that represents the progression through the cycle. Aside from parameterizing the gait, ideally the phase variable is invariant across different subjects and does not depend on parameters such as the person's mass or height \cite{Villarreal2017}. In \cite{gregg2014}, the heel-to-toe movement of the Center of Pressure (CoP) served as the phase variable for determining progression through the stance phase, whereas the swing phase was controlled by two impedance-based states. In \cite{gregg2014b}, the authors investigated additional phase variables for locomotion and found that the global thigh angle is a suitable piecewise monotonic signal that can be used to control the stance and swing phases separately. By also using the integral of the global thigh angle, the phase variable was made continuous across the gait cycle and was implemented in a powered knee-ankle prosthesis for use by amputee subjects \cite{Quintero2018}.

Everyday tasks comprise both rhythmic activities, such as walking, as well as non-rhythmic activities, such as stepping over obstacles. A controller strictly based on behavior in a rhythmic task, such as the unified controller presented in \cite{Quintero2018}, will encounter problems for non-rhythmic motions and thus will not be practical for real-world use. Previous studies such as \cite{ha2011,hargrove2011,Hoover2013} attempted to enhance volitional control of rhythmic and non-rhythmic motions using information contained in the bioelectric signals of the residual limb, such as those acquired from electromyography (EMG). However, EMG quality is  highly dependent on physical factors such as electrode placement, movement artifact, and electromagnetic noise; physiological factors such as muscle and nerve fatigue; and anatomical factors such as volume conduction, which causes ``mixing'' of signals from different underlying  muscles when using surface electrodes. In the case of transfemoral amputees, the muscles used for controlling the ankle joint do not exist. The information may still be recovered if a nerve re-innervation procedure was carried out during amputation, but sophisticated ``unmixing'' algorithms have to be used to decipher individual muscle activity from the EMG \cite{Hargrove2013}. In light of this knowledge, we seek a more reliable solution using only mechanical measurements.  

As a first attempt for such a control scheme, Villarreal et al. used the thigh angle and a stance/swing detection switch to implement a piecewise phase variable for non-rhythmic control \cite{Villarreal2017b}. However, the controller was problematic during transitions, as using solely the foot contact condition for transitioning between stance and swing phase variables would result in jumps and oscillations. To avoid such jumps, pushoff was eliminated, but it consequently made walking at greater speeds difficult and inefficient. Moreover, the undesired jumps would still occur when standing, as the subject shifted their weight to the sound leg. This motivated the design of a controller based on a Finite State Machine (FSM) and preliminary experiments with an amputee subject demonstrated its functionality for different non-rhythmic and rhythmic tasks  \cite{rezazadeh2018}. 

In this article, we extend this investigation by showing how the proposed controller can improve the performance compared to passive transfemoral prostheses. This assessment includes both rhythmic and non-rhythmic tasks (which was not possible using our previous controller \cite{Quintero2018}). Comparisons of walking performance with passive and powered prostheses have been done in works such as \cite{Armannsdottir2017} (for a powered ankle) and \cite{Jayaraman2018} (for a powered knee-ankle). However, both these works rely on several sessions of training in order to obtain meaningful improvements with the powered prostheses. Specifically, in \cite{Armannsdottir2017}, it was shown that without training, the outcomes may only have slight improvements. In contrast, we show that using the presented controller, clinically significant improvements can be observed immediately after a brief tuning/acclimation session (about 10 minutes) with the powered leg.

In \cite{Jayaraman2018}, the positive effects of a powered leg on back muscles and gait energy expenditure were investigated. Such symptoms often originate from compensations that amputees adopt to overcome the lack of power in passive prostheses \cite{waters1999energy,gailey2008review,kulkarni1998association,Devan2014}. In this work, we study the kinematic and kinetic attributes characterizing these compensations. The most common compensations associated with the use of passive prostheses are related to prosthetic foot clearance resulting from under-powered plantarflexion during pushoff and knee flexion during swing \cite{atlas2004compens,Drevelle2014}. Since individual amputees develop different compensatory mechanisms, we focused our analyses on traits that were the most apparent with the amputee participant, namely, vaulting and hip circumduction. Vaulting emerges as excessive plantarflexion during midstance of the sound leg, whereas hip circumduction is characterized by excessive hip abduction of the prosthetic side during swing, resulting in lateral deviation of the prosthetic foot. We hypothesized our powered leg would help mitigate these compensatory mechanisms by providing active plantarflexion and knee flexion to better aid the foot clearance.

In both \cite{Armannsdottir2017} and \cite{Jayaraman2018} the studies were limited to comparison of powered and passive legs for rhythmic (treadmill walking) tasks. A direct outcome of the present phase-based controller is its additional ability to enable a range of non-rhythmic tasks. Based on this, we studied some common non-rhythmic tasks that amputees may face in daily life. For backward walking, as a first task tested for this purpose, two different gait patterns are recognizable. In a ``step-to'' pattern, the sound leg leads the motion while the prosthetic leg follows, seldom passing beyond the sound leg. In contrast, in a ``step-through'' pattern, both sound and prosthetic legs alternately lead the motion and the swing leg  passes the stance leg. We analyze these gait patterns during backward walking trials to determine how the powered leg performs compared to the passive leg.

The next non-rhythmic task we tested was stepping over an obstacle. We predicted the subject would be able to successfully maneuver his intact joints into a configuration that maintains the prosthetic knee and ankle in a sufficiently flexed position while guiding his hip joint over the obstacle. This kind of maneuver is quite difficult or even impossible with passive prostheses or the state-of-the-art powered prostheses designed for walking. Another task that we tested was kicking a soccer ball. Kicking is a non-rhythmic movement that requires powerful but controlled extension of the knee. Functional abilities are compared between the powered and passive legs during these tasks.

The paper is organized as follows. In \Cref{sec:ControlDesign}, we describe the new phase variable, its states and corresponding transitions, and the implementation of the controller. \Cref{sec:Experiments} addresses the experimental setup, protocol, and methods. Next,  in \Cref{sec:Results}, the results of the experiments are presented. Based on that, in \Cref{sec:Discussion}, the obtained results are discussed and the performance of the controller for different tasks is analyzed. At the end, we present a series of conclusions together with suggestions for future works.


\section{Control Design}
\label{sec:ControlDesign}
This section presents the design of the proposed scheme for non-rhythmic and rhythmic control of a powered knee-ankle prosthesis. First, we explain the use of virtual constraints for formulating the desired knee and ankle joint trajectories. Next, we describe the design of our proposed phase variable for parameterizing the virtual constraints in different stages of the gait cycle. Finally, we discuss how the controller is implemented on a powered prosthesis.

\subsection{Virtual Constraints}
Virtual constraints, as introduced in \cite{Westervelt2004, Westervelt2007}, are a useful tool to represent time-invariant trajectories, which can considerably simplify the process of controlling periodic orbits. Originally, virtual constraints were introduced as relationships among generalized positions (angles), which is analogous to a holonomic set of kinematic constraints. More recently, nonholonomic virtual constraints have also been used in legged robots applications \cite{Rezazadeh2015, Griffin2017}. Generally, virtual constraints define the desired trajectories for the controlled degrees of freedom in the following form:
\begin{equation}
\label{eq:virtualConstraint}
q_i^d=h(s),
\end{equation}
\noindent where $s$ is a monotonic function of positions (for holonomic virtual constraints), or positions and velocities (for nonholonomic virtual constraints), and is usually scaled between 0 and 1. 

In legged robot applications, $s$ is normally reset every step, and continuity is preserved by imposing equality conditions on $h$ and $\partial h/\partial q$ at $s=0$ and $s=1$. This is a convenient choice for legged robots, especially considering there are sensors on both legs for computing the phase variable. For a prosthetic leg application, in order to avoid attaching sensors on the sound leg of the subject,  it is desirable to use only onboard sensors from the prosthesis. This is equivalent to resetting the phase variable at the end of each stride, rather than each step. In this case, \eqref{eq:virtualConstraint} represents the desired, periodic trajectories for the entire stride \cite{quintero2018toward}.

Due to their dependence on velocities or integrals, nonholonomic virtual constraints are sensitive to changes in speed and are thus not suitable for a controller that is intended to work in a wide range of non-steady activities. A good example is the integral-based unified controller presented in \cite{Quintero2018}, which worked well in normal-speed steady-state walking, but was unreliable for slow speeds and was unable to perform non-rhythmic motions. Therefore, we establish our control scheme on a holonomic phase variable in order to make it speed-independent. Motivated by the results of \cite{gregg2014b} and \cite{Villarreal2017}, and since the holonomicity of the thigh angle makes it an ideal selection for a volitional controller, we use this variable as the basis for our controller. In what follows we show how this angle is used to construct our phase variable.

\subsection{Constructing the Phase Variable}
\label{section:PhaseVariable}
\begin{figure}
	\centering
	\includegraphics[width=0.9\linewidth]{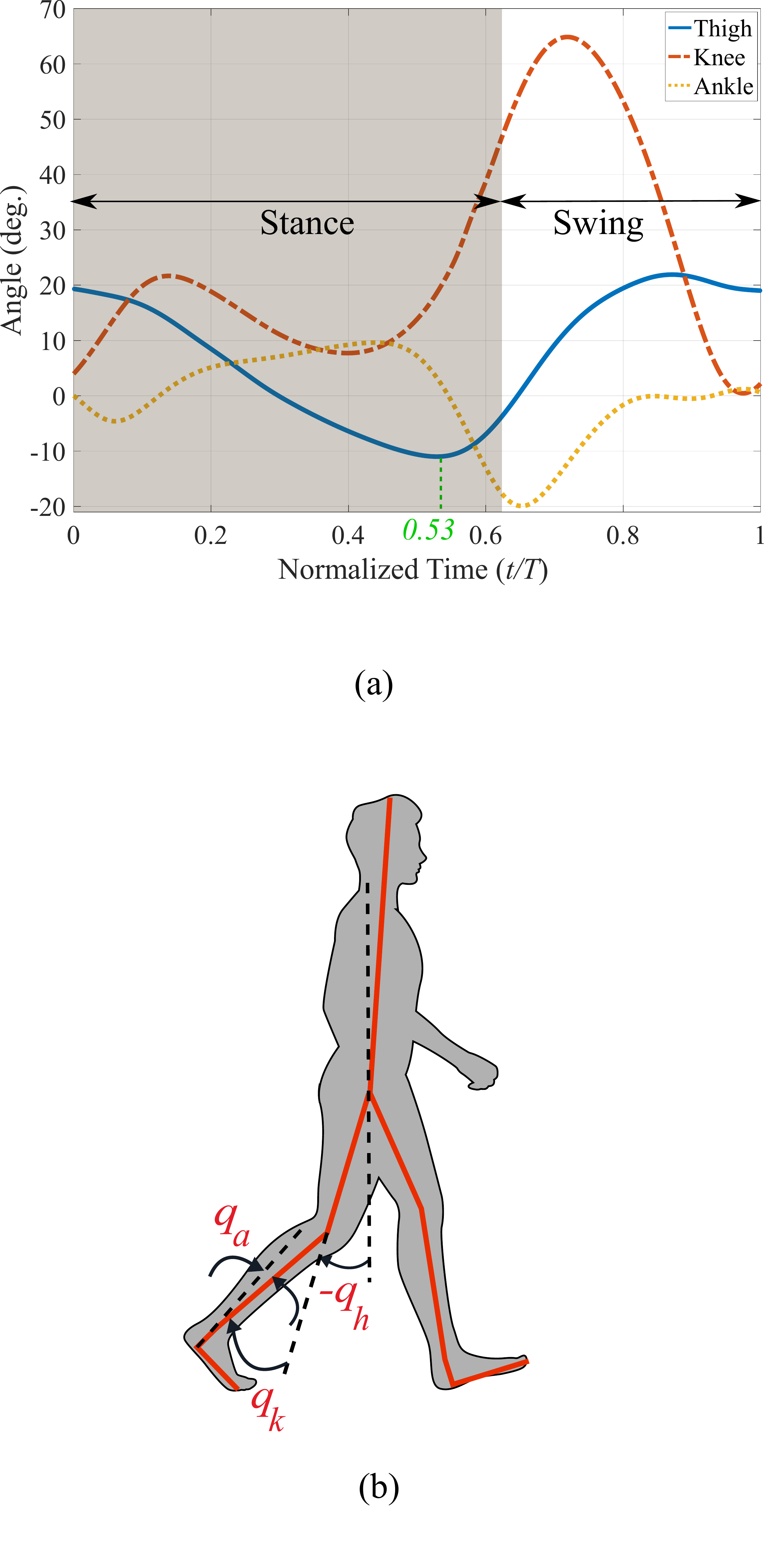}
	\caption{(a) Human leg's joint angle trajectories during one stride of walking with normal speed and stride period $T$ \cite{Winter2005}. (b) Definition of the joint angles.}
	\label{fig:humanAngles}
\end{figure}

As mentioned before, we aim to use thigh angle $q_h$ (Fig. \ref{fig:humanAngles}(b)), for defining our holonomic phase variable. In what follows, we will show how this variable can be used for this purpose and what other measurements are necessary.

Fig. \ref{fig:humanAngles}(a) depicts the thigh, knee, and ankle angle trajectories during one stride of a normal able-bodied walking gait on flat ground \cite{Winter2005}. Note that the thigh angle is not a monotonic signal throughout the stride. As a result, each value of $q_h$ corresponds to at least two points in the cycle (one in the descending part of $q_h$ and one in the ascending part), making the determination of a unique $s$ based solely on $q_h$ impossible. To avoid this problem, and also to keep the benefits of a holonomic system, we propose to use a set of \emph{piecewise holonomic} virtual constraints. The idea is to divide the gait cycle into different sections, where each section corresponds to a monotonic (either ascending or descending) thigh angle trajectory.

From Fig. \ref{fig:humanAngles}, the thigh angle trajectory during a stride can roughly be divided into two monotonic sections (neglecting the small retraction section at the end); it is descending after heel strike ($t/T= 0$) and through the stance phase until the trajectory reaches its minimum at $t/T= 0.53$, and then becomes ascending. Note that the swing phase starts a little later, at $t/T= 0.63$.  An obvious way to transitioning between these two states is using the sign change of the thigh angle's rate, $\dot{q}_h$. In practice this proves to be a very sensitive signal, because velocities can change rapidly, which results in large discontinuities in the virtual constraints and in undesirable transitions. For this reason and since these two monotonic sections approximately correspond to stance and swing phases, in \cite{Villarreal2017b} a foot contact sensor was used for transitioning between these two states. The first problem with this approach is that the minimum thigh angle does not exactly correspond to the foot takeoff ($t/T= 0.53$ versus $t/T= 0.63$) and thus part of pushoff will be performed when the leg is already in swing. Moreover, this approach assumes that the thigh angle exactly follows the reference trajectory. If the minimum thigh angle is larger than the reference trajectory's minimum (shorter step), there will be a jump in the virtual constraints. Conversely, if the minimum thigh angle is less than the reference (longer step), the virtual constraint will saturate, which leaves pushoff half-completed. These undesirable features can be seen in the results of \cite{Villarreal2017b}.

\begin{figure*}
	\centering
	\subfigure[]{\includegraphics[width=.46\linewidth,trim={5cm 3.1cm 5cm 5cm},clip]{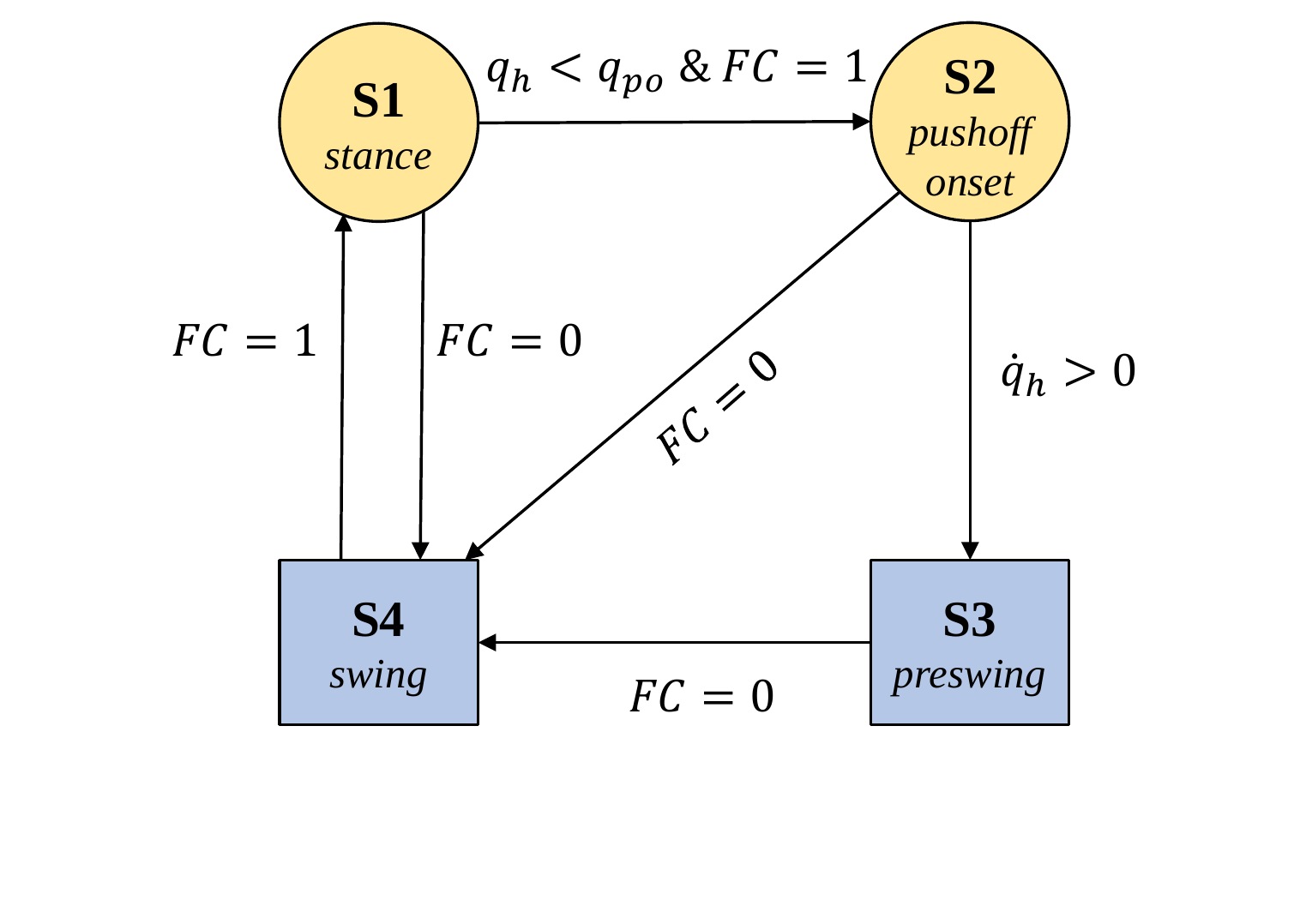}\label{fig:FSMforward}}
	\subfigure[]{\includegraphics[width=.53\linewidth,trim={3.8cm 4cm 3.5cm 2cm},clip]{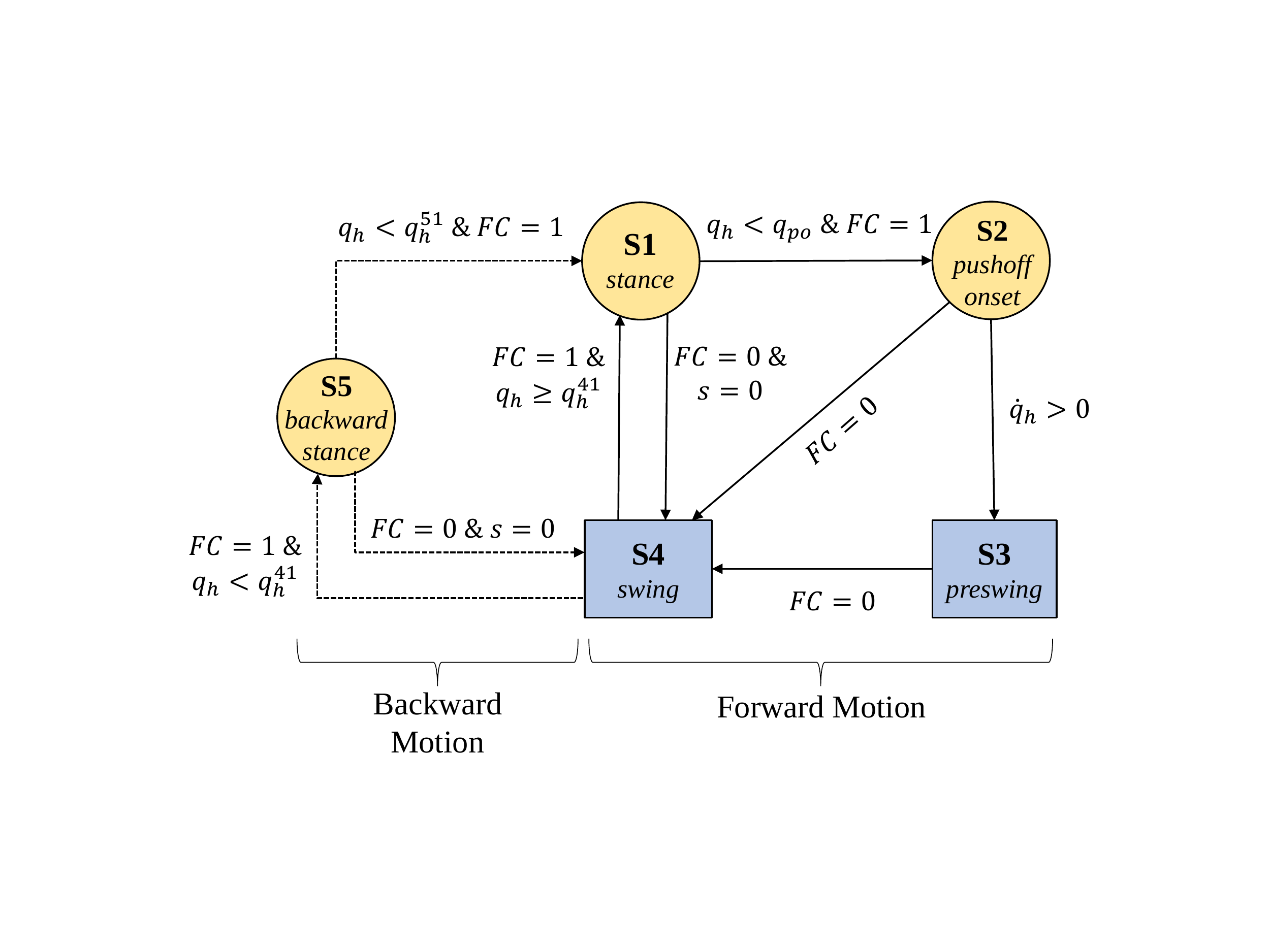}\label{fig:FSMcomplete}}
	\caption{
		\subref{fig:FSMforward} The preliminary finite state machine based on forward walking. The phase variable, $s$, in the yellow circle states is obtained from \eqref{eq:stancePV}, and in the blue rectangle states from \eqref{eq:swingPV}.
		\subref{fig:FSMcomplete} The complete finite state machine for computing the phase variable for control of the prosthetic leg for forward and backward walking and general non-rhythmic tasks. Again, the yellow circles correspond to \eqref{eq:stancePV}, and the blue rectangles to \eqref{eq:swingPV}.}
	\label{fig:FSM}
\end{figure*}

To resolve these problems, we propose to have two supplementary states (in addition to stance and swing) to represent pushoff. The result is depicted in Fig. \ref{fig:FSMforward} in the form of an FSM with four states, where S1 and S2 pertain to the descending part of the thigh trajectory, and S3 and S4 correspond to the ascending part. Note that S1, S2, and S3 are all parts of the stance phase, and thus for all of these states $FC=1$ ($FC$ represents foot contact as a binary signal). For this reason, we use other variables to define these transition conditions. Namely, transitioning from S1 (stance) to S2 (pushoff onset) occurs at a specific thigh angle ($q_h=q_{po}$), and transitioning from S2 to S3 (pre-swing) occurs when $\dot{q}_h=0$. The tunable constant $q_{po}$ represents the thigh angle at the start of pushoff and its default value is obtained from the thigh angle at the maximum ankle angle in the reference trajectory (Fig. \ref{fig:humanAngles}, from which $q_{po}=-8.4^{\circ}$). As previously mentioned, a transition based on velocity is accompanied with the risk of sensitivity and sudden jumps in virtual constraints. Although these jumps would be small due to the small range of thigh angles represented by S2 and S3, we propose a two-step approach to completely eliminate such discontinuities. In the first step, the transitions from S1 to S2 and from S2 to S3 are designed to be unidirectional, resulting in only one possible jump from S2 to S3. To eliminate this single jump, in the second step we reset the associated parameters based on the information from the sensors. This will be explained in the definition of $s$ in what follows.

For S1 and S2, the phase variable can be computed from a shift and scale of the thigh angle:
\begin{equation}
\label{eq:stancePV}
s=\frac{q_h^{0}-q_h}{q_h^{0}-q_h^{min}}\cdot c,
\end{equation}
\noindent where $q_h^{0}$ and $q_h^{min}$ are constants whose default values are touchdown value and the minimum of the reference thigh angle trajectory, respectively. These two parameters can be tuned if the subject prefers a different step length. The constant $c$ is also tunable and is related to the ratio of the stance phase to the whole cycle. The default value of $c$ is the normalized time at which $q_h$ reaches its minimum, which is 0.53 in Fig. \ref{fig:humanAngles}.

Since the transitioning from S2 to S3 is based on the change of sign of $\dot{q}_h$, S3 pertains to the ascending part of the thigh angle. To form a continuous phase variable and to avoid jumps at each transitioning from S2 to S3, we record the values for $s$ and $q_h$ and name them $s_m$ and $q_{h,m}$, respectively. The phase variable in preswing (S3) and swing (S4) phases is then computed from
\begin{equation}
\label{eq:swingPV}
s=1+\frac{1-s_m}{q_h^0-q_{h,m}}\cdot (q_h-q_h^0).
\end{equation}
\noindent Note that $s=s_m$ at $q_h=q_{h,m}$, and $s=1$ at $q_h=q_h^0$. For both \eqref{eq:stancePV} and \eqref{eq:swingPV}, the phase variable is saturated between 0 and 1.

An additional factor to consider for the preswing phase is the tendency of the leg to oscillate as the load is removed from it. This is eliminated by imposing a unidirectional filter on the phase variable in S3. That is, in the discrete time instance $k$:
\begin{eqnarray}
s(k)\geq s(k-1), && \text{when }S=S3,
\end{eqnarray}
\noindent where $S$ is the current state. Note that this condition is not required for S2, as it transitions to S3 at the first instance when $\dot{q}_h$ (and hence $\dot{s}$) crosses zero.

The FSM of Fig. \ref{fig:FSMforward} together with the phase variable definition in \eqref{eq:stancePV} and \eqref{eq:swingPV} constitute a control paradigm based on a forward walking scheme. However, a volitional controller needs to also manage situations in which the motion is interrupted or even reversed. Due to the holonomic nature of the designed phase variable, it is invariant to the direction of motion. Therefore, the problem can only arise during transitions. The most critical situation happens when the leg is in swing and it touches the ground behind the body (backward walking). According to Fig. \ref{fig:FSMforward} the state transitions to S1 and then immediately to S2 (and perhaps S3), which leads to pushoff and does not allow the subject to put weight on the leg. In order to avoid this, we added another state, S5, to the FSM (Fig. \ref{fig:FSMcomplete}). This new state keeps the leg in stance phase when walking backward, and it transitions to pushoff only if the subject resumes moving forward. With this new state, we define the transitions for our controller as follows:

\begin{enumerate}
	\item \textit{Transition from S1 or S5 to S4:} The primary condition for transitioning between stance and swing is foot contact. However, in conditions such as standing still, if the leg is unloaded for a moment (i.e., shifting weight to the sound leg), a transition to swing can result in a sudden and undesirable flexion of the knee. To avoid this, we require that the transition to swing happens either after pushoff (i.e., through S2 and S3), or directly from S1 or S5 to S4 at maximum thigh angle ($s=0$). Obviously, for transitioning from S5 to pushoff, the state first needs to go to S1, as discussed next.
	\item \textit{Transition from S5 to S1:} This transition happens when the subject steps backward and then decides to move forward. The transition condition is given by $q_h<q_h^{51}$, where $q_h^{51}$ is a tunable constant. Note that $q_h^{51}>q_{po}$, in order to avoid direct transitioning to pushoff. For our experiments, we set $q_h^{51}=-6^{\circ}$.
	\item \textit{Transition from S4 to S1 or S5:} Since stance is a more reliable state for the subjects (they can put their weight on the leg), the condition for transitioning from S4 to S1 or S5 is less strict compared to S1 and S5 to S4. When foot contact happens ($FC=1$), the transition will be to S1 if $q_h\geq q_h^{41}$, otherwise it will be to S5, where $q_h^{41}$ is a tunable constant. For our experiments, we set $q_h^{41}$ to zero, which is equivalent to transition from S4 to S1 for a forward step or to S5 for a backward step.
	
\end{enumerate}

\noindent Fig. \ref{fig:FSMcomplete} summarizes the states and the corresponding transitions.

\subsection{Control Design Based on Virtual Constraints}
Having computed the phase variable, the next step is obtaining the virtual constraints. In this work, we follow the approach of \cite{Quintero2018} in using Discrete Fourier Transform (DFT) in order to generate virtual constraints for knee and ankle, based on data from normal-speed flat-ground human walking provided in \cite{Winter2005}. In this form, the desired joint angles can be computed as

\begin{equation}
\label{eq:VCDFT}
\begin{aligned}
q_i^d=h(s)= & \frac{1}{2}\rho_0+\frac{1}{2}\rho_{_{N/2}}\cos(\pi Ns) \\
&+\sum_{k=1}^{N/2-1} \left[\rho_k\cos(\Omega_ks)-\psi_k\sin(\Omega_ks)\right],
\end{aligned}
\end{equation}

\noindent where $\rho_k$ and $\psi_k$ are the coefficients of real and imaginary parts of DFT, respectively.

In the next step, the desired knee and ankle angle obtained from \eqref{eq:VCDFT}, are imposed using a Proportional-Derivative (PD) position controller. Noting that the position error for joint $i$ is
\begin{equation}
e_i=h(s)-q_i,
\end{equation}
\noindent the commanded motor torque is obtained from
\begin{equation}
\tau_i=K_{p,i}e_i+K_{d,i}\dot{e}_i,
\end{equation}
\noindent where $K_{p,i}>0$ and $K_{d,i}>0$ are PD control gains for joint $i$.

Fig. \ref{fig:BlockDiagram} displays the block diagram of the proposed controller.


\section{Experimental Methods}
\label{sec:Experiments}

\subsection{Materials}

\begin{figure*}
	\centering
	\subfigure[]{\includegraphics[width=0.84\linewidth,trim={0.15cm 5.0cm 0cm 4.5cm},clip]{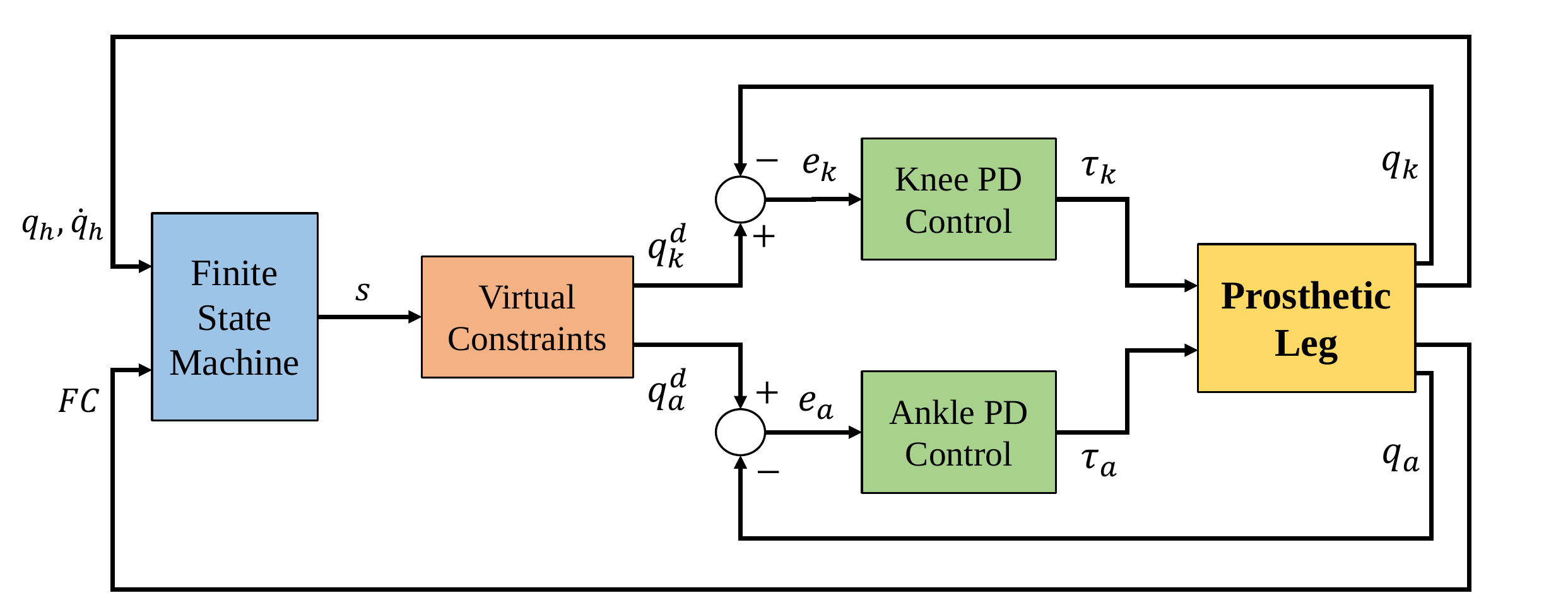}\label{fig:BlockDiagram}}
	\subfigure[]{\includegraphics[width=.6\linewidth,trim={0cm 0cm 0cm -2cm},clip]{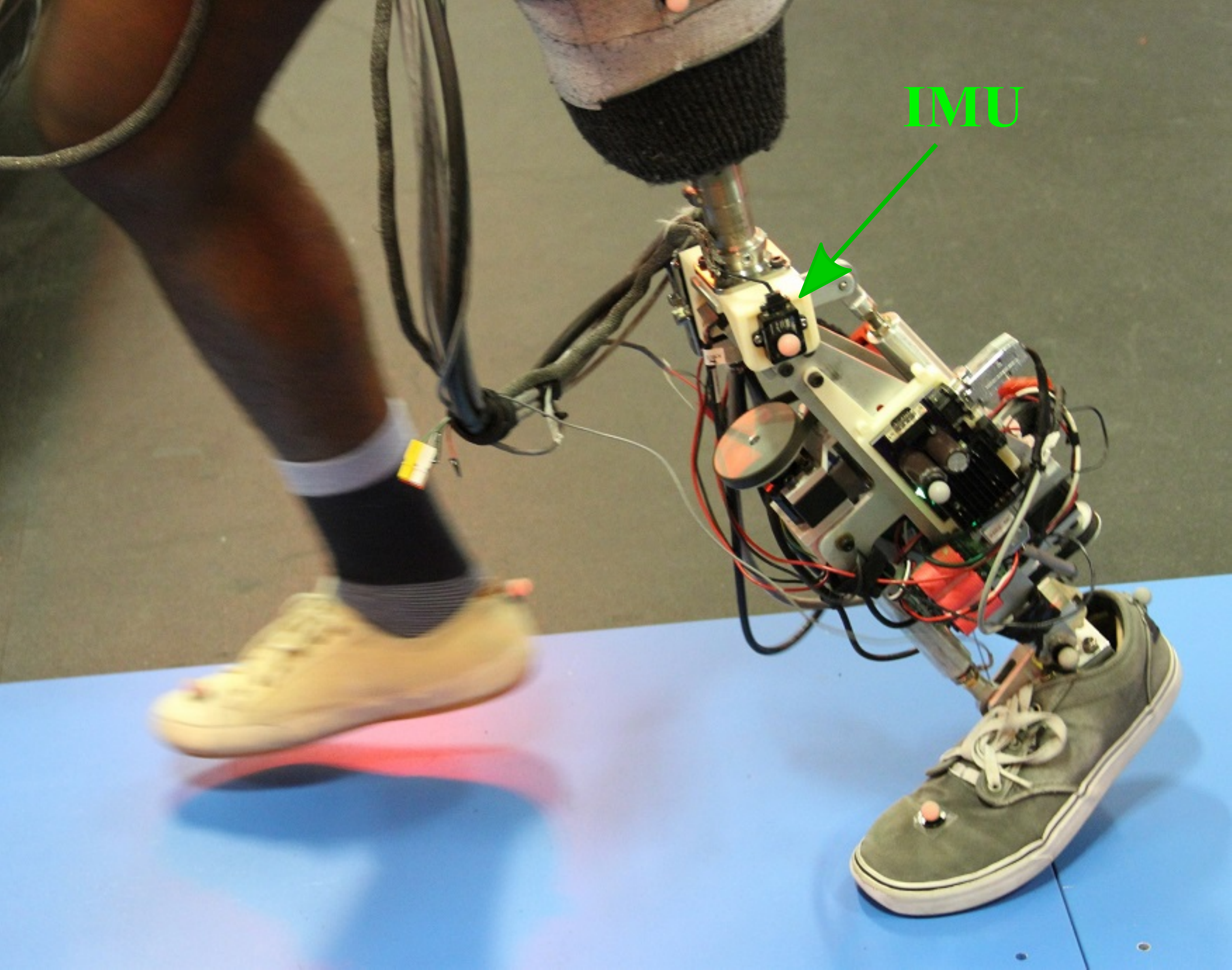}\label{fig:legPhoto}}
	\caption{\subref{fig:BlockDiagram} Block diagram of the proposed controller for the knee-ankle prosthesis. $q_k$ and $q_a$ represent knee and ankle joint angles, respectively, and $q_k^d$ and $q_a^d$ are their desired values.\subref{fig:legPhoto} The powered knee-ankle prosthetic leg worn by the transfemoral amputee participant.}
	\label{fig:hardware}
\end{figure*}

The knee-ankle powered leg used for our experiments is shown in Fig. \ref{fig:legPhoto}. Each joint is equipped with a Maxon EC-4pole 30, 200 Watt, three-phase brushless DC motor driving the joints through a timing belt and a Nook 2-mm lead ball screw. Due to greater torques in the ankle joint, the timing belt ratio for the ankle is twice that for the knee.

The joints and motors are equipped with optical encoders (Maxon 2RMHF for motors and US Digital EC35 for joints). An IMU sensor (LORD MicroStrain, 3DMGX4-25) is used to measure the global thigh angle, as shown in Fig. \ref{fig:legPhoto}. Foot contact condition is determined using a force sensitive resistor sensor (FSR - FlexiForce A401, Tekscan Inc.) located inside the pyramid adapter of the prosthetic foot. 

The computation and control is done offboard via a tethered connection to a dSPACE DS1007 system with Freescale OorIQ P5020, dual-core, 2 GHz PowerPC processor. The commanded torques from the computer are sent to an Elmo Gold Twitter R80/80 amplifier, which controls the motors. The powers for the motors and the FSR are provided through DC power supplies (Agilent Technologies 6673A for the motors and BK Precision 1761 for the FSR). See \cite{Quintero2018} for further details on the design of the prosthetic leg and the hardware specifications.

Motion capture data was collected using a 10-camera VICON system. A 16-marker lower body model was used (Plug-in-Gait - VICON). Standard data processing procedures, including gap filling and filtering (fourth-order 6 Hz cut-off low-pass Butterworth), were applied. A walking platform with a single 3-axial force plate (Kistler) positioned midway of the walkway length was constructed. Handrails were provided alongside the entire length of the platform. A non-instrumented treadmill with a safety harness was used for collection of continuous gait data.

\subsection{Procedure} \label{subsection_procedure}

The controller was first tested and tuned through a set of walking experiments on a flat-ground treadmill with an able-bodied subject using the powered leg with a bypass system \cite{quintero2016preliminary}. Any tunable parameter discussed in Section \ref{sec:ControlDesign} was appropriately changed until the subject was able to comfortably walk in different speeds. The joint trajectories for the trials with three different speeds (slow: 0.7 m/s; normal: 0.9 m/s; and fast: 1.1 m/s) were recorded for qualitative comparison with those of the amputee subject. After this set of tests, the main experiments were conducted with an amputee subject with the same set of parameters and with no change. The amputee participant was a 32-year old male with a height of 1.75 m, weight of 76 kg with the passive leg, and weight of 78.9 kg with the powered leg. His left leg was amputated 11 years before these experiments and he used an Ottobock 3R60 knee together with an Ability Rush ankle-foot prosthesis as his everyday passive leg. The participant had no prior experience with the powered prosthesis device being tested, and he had no neuro-muscular disorder that is known to affect gait, balance, or muscle activity. The experimental protocol was reviewed and approved by the Institutional Review Board (IRB) at the University of Texas at Dallas. 

Prior to the experiment with the amputee subject, standard procedures for fitting and tuning of the prosthetic leg were carried out by a licensed Prosthetist. For fitting the powered prosthesis, the participant's daily-use socket was used. The experiments were designed in two different sets, namely non-rhythmic and rhythmic. The non-rhythmic experiments were done on the walkway platform, while the rhythmic experiments were conducted on the treadmill. The subject first performed both sets of tasks with his passive prosthesis, and then he repeated them with the powered leg. As mentioned, for all these tasks, the control parameters introduced in \Cref{sec:ControlDesign} remained the same.

As the first overground experiment, the subject was asked to start from rest, walk forward between the handrails with a comfortable speed, and stop at the end of the walkway. The experiment was repeated until we captured three clean foot strikes on the force plate for each leg (prosthetic, sound). After that, we examined the invariance of the controller to the direction of locomotion and the ability of reversing the direction of walking at the subject's will. For this purpose, we asked the participant to walk backward on the walkway, as well as a combination of forward and backward transitions.

After that, we tested the ability of the subject to step over an obstacle, specifically an 85-mm high wooden block. The goal was to step over it first using the sound leg and then the prosthesis. No restrictions were placed on the number of practice trials. No additional guidance was provided regarding the ideal foot placement or maneuvering strategy for crossing the obstacle; the participant was encouraged to explore and practice his preferred strategy. After this experiment, the subject was asked to kick a soccer ball to demonstrate the fast extension of the powered knee following a quick forward motion of the hip in an activity other than walking. This test concluded the overground experiments.

Flat-ground treadmill trials for rhythmic tasks were performed at three speeds: slow, normal, and fast. First, the participant was instructed to adjust the treadmill speed to his self selected speed (i.e., normal speed) with his passive prosthesis. The slow and fast speeds were then taken as the minimum and maximum speeds, respectively, at which the participant was able to maintain a stable and comfortable gait with his passive leg. Data were captured for 60, 60, and 45 seconds for slow, normal, and fast speeds, respectively, where recording was only started after the participant was able to produce a reasonably consistent gait cycle (as visually inspected). With the powered leg, the subject was able to reach higher speeds than the maximum speed with the passive leg. Thus, in addition to the previous speeds, we tested and recorded the maximum speed at which the subject was able to comfortably walk with the powered leg for the duration of 30 seconds.

\subsection{Data and Statistical Analyses} \label{subsection_data_analysis}
For comparing the joint angle trajectories of the prosthetic legs with those of the reference able-bodied subjects (adopted from \cite{Winter2005}), we computed the Pearson correlation coefficient for each case, as in\cite{Jayaraman2018}. In the ideal case (trajectories identical to reference able-bodied trajectories), this coefficient would be equal to 1.

Vaulting is usually quantified by measuring peak ankle flexion power during single support \cite{Drevelle2014}. However, in the present work, we utilize an alternative to kinetic quantification, using only kinematic parameters. As it will be shown in the next section, this approach measures the peak sagittal-plane global foot angle (with respect to the ground) during single-support, which corresponds to the point of zero velocity.

Similar to \cite{Awad2017}, we quantified hip circumduction as the medio-lateral range of motion of the prosthetic ankle marker coordinates during a stride. This method corresponds better with what a clinician observes visually than other measures such as hip frontal-plane range of motion \cite{Kerrigan2000}.

The symmetry index ($SI$) between the sound and prosthetic sides was quantified as 
\begin{equation}
SI = \left\vert\frac{V_{Prosth}-V_{Sound}}{\frac{1}{2}(V_{Prosth}+V_{Sound})}\right\vert,
\end{equation}
where $V_{Prosth}$ and $V_{Sound}$ are the gait variables corresponding to each side \cite{patterson2010evaluation}. Based on this, $SI=0$ represents perfect symmetry and increasing deviation from zero corresponds to increasing asymmetry.

For backwards walking, we assessed the improvement in gait type from ``step-to'' gait (observed with the passive leg) to a ``step-through'' gait (observed with the powered leg) \cite{Thrasher2006}. This asymmetry in backwards motion cannot be captured using standard step length symmetry assessments which are based on the distance covered by each foot with respect to itself. Instead, we assessed the distance along the direction of walking between the prosthetic versus sound ankle marker coordinates at their respective mid-stance positions. This allowed a relative (prosthetic versus sound) measure of foot positions.  

The ability to cross over the obstacle was quantified by the height of the toe marker above the obstacle and also qualitatively assessed by plotting toe marker trajectories over the obstacle. Kicking the soccer ball was quantified by measuring ball velocity (estimated using coordinates of reflective markers affixed to the ball).  

Means and standard deviations were calculated for the biomechanical variables discussed in the previous section for each applicable condition resulting from a combination of 1) prosthetic device (powered, passive), 2) side (prosthetic, sound), and 3) speed (fast, normal, slow).


\section{Results}
\label{sec:Results}
A supplemental video of these experiments is available for download. The main goal of this work was to demonstrate the ability of the controller to facilitate both rhythmic (constant-speed walking) tasks as well as a range of non-rhythmic (i.e., volitional) tasks and the improvements compared to the amputee subject's passive leg. Our participant was able to successfully complete all (rhythmic and non-rhythmic) tasks that were part of the testing protocol as described in \Cref{subsection_procedure}. We present both qualitative as well as quantitative results in what follows.

\subsection{Non-Rhythmic Tasks}
\begin{figure}
	\centering
	\subfigure[]{\includegraphics[width=1\linewidth,trim={1cm 1.5cm 2cm .5cm},clip]{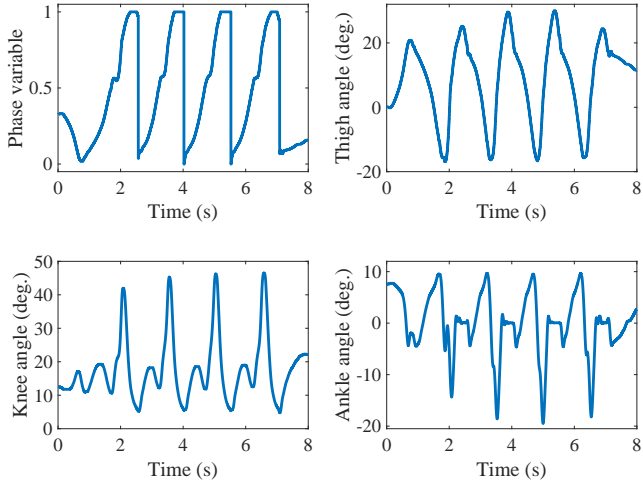}\label{fig:handrailFWD}}
	\subfigure[]{\includegraphics[width=1\linewidth,trim={1cm 1.5cm 2cm .5cm},clip]{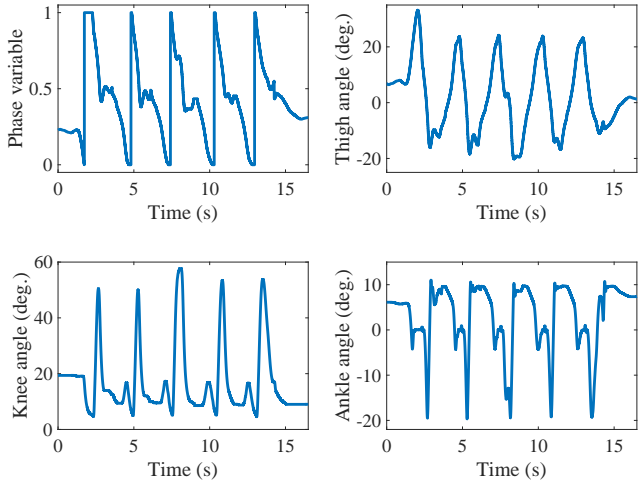}\label{fig:handrailBWD}}
	\caption{Phase variable and joint angle plots for \subref{fig:handrailFWD} a forward walking trial between handrails, and \subref{fig:handrailBWD} a backward walking trial.}
	\label{fig:handrails}
\end{figure}

The first non-rhythmic tasks were overground forward and backward walking between handrails, including start and stop. Fig. \ref{fig:handrailFWD} displays the phase variable and joint angles through an overground forward walking trial. The subject started from rest (almost vertical leg), walked across the walkway, and stopped at the end. The change of the minimum ankle angle across strides is particularly interesting, as it represents the extent of pushoff. As the subject started from rest and increased his walking speed, the ankle plantarflexion also increased (i.e., larger pushoff) until the last stride where the subject decreased his speed and pushoff became smaller correspondingly.

\begin{figure}
	\centering
	\includegraphics[width=0.7\linewidth]{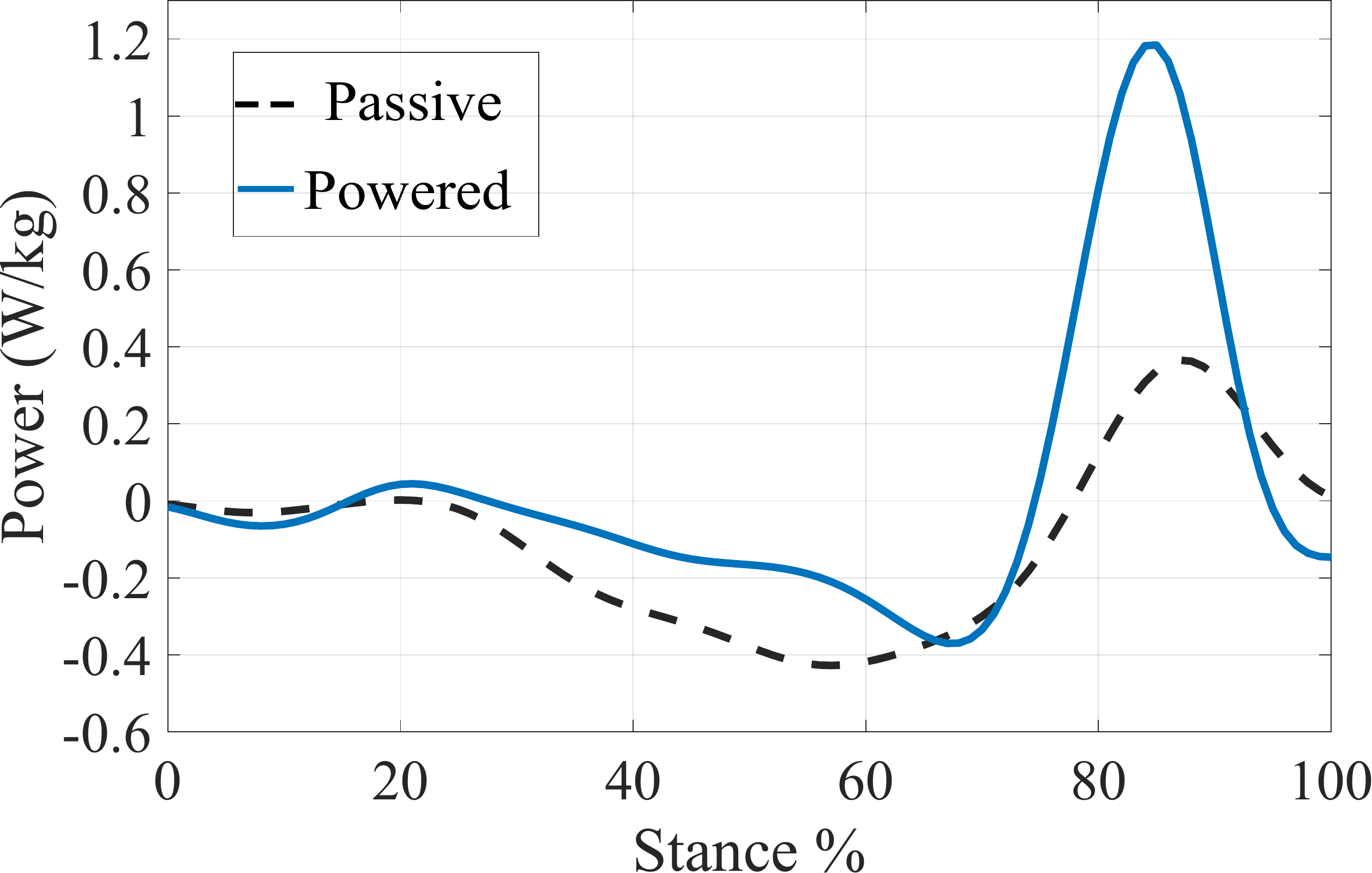}
	\caption{Comparison of joint ankle powers during stance phase for overground trials with powered and passive legs. A significantly higher pushoff power is observed for the powered leg compared to the passive leg.}
	\label{fig:anklePower}
\end{figure}

The representative joint ankle powers for overground walking with powered and passive legs are displayed in Fig. \ref{fig:anklePower}. Although the passive ankle cannot produce power, its compliance allows for storage and release of energy through the stance phase. The comparison of the pushoff powers shows that the peak generated by the powered leg is more than three times larger compared to that of the passive leg. We will discuss the improvements in gait characteristics associated with this increase in the next subsection.

The results of a representative backward walking trial with the powered leg are depicted in Fig. \ref{fig:handrailBWD}. The holonomic nature of the controller enabled the subject to comfortably reverse his direction of motion and still maintain a smooth gait. Note that the phase variable has a reverse trajectory compared to Fig. \ref{fig:handrailFWD}. Because of the natural limitation of the passive prosthesis, the participant was forced to adopt a step-to pattern of walking, whereas he was able to walk using a normal step-through pattern using the powered leg. Our results show that use of the powered leg substantially improved the symmetry index for backwards walking: from 0.56 (passive leg) to 0.26 (powered leg). See Table \ref{tab:backward_step} for details.

\begin{table}
	\caption{Backward walking step symmetry: passive (Pas) versus powered (Pwr), and sound (S) versus prosthetic (P).}
	\label{tab:backward_step}
	\begin{center}
		\renewcommand{\arraystretch}{1.7} 
		\begin{tabular}{|c||c|c|c|c|} \hline
			& \textbf{Pas P}   & \textbf{Pas S}   & \textbf{Pwr P}  & \textbf{Pwr S}\\ \hline
			\textbf{Length (mm)}           & 281 (32)     & 498 (44)     & 669 (44)   & 513 (30) \\ \hline
			\textbf{SI}       &	\multicolumn{2}{c|}{0.56}	        & \multicolumn{2}{c|}{0.26}	\\ \hline	 
			
		\end{tabular}
		\renewcommand{\arraystretch}{1}
	\end{center}
\end{table}

\begin{figure}
	\centering
	\includegraphics [width=1\linewidth]{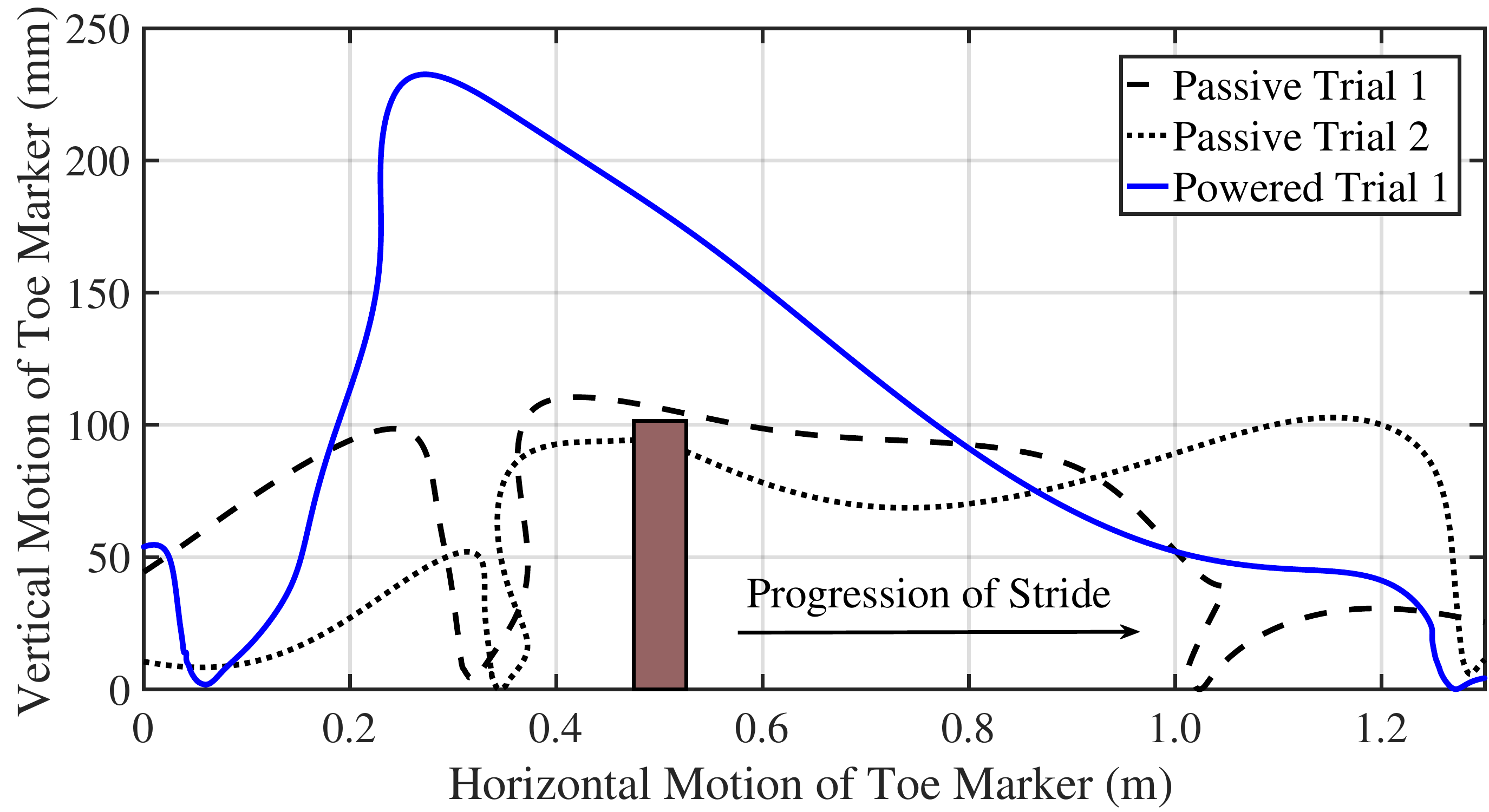}
	\caption{Illustration of obstacle crossing using toe marker trajectory. A high clearance can be seen using the powered leg, whereas much lower clearances are seen using the passive leg. For better illustration, different scales for horizontal and vertical axes have been applied.}\label{fig:obstacle}
\end{figure}

Crossing over an obstacle was another non-rhythmic everyday task that we tested and compared between the passive and powered legs. Fig. \ref{fig:obstacle} shows the planar path of the toe marker and the obstacle. It can be seen that the powered leg (blue solid curve) provided substantially greater toe clearance compared to the passive leg (dashed curve). In fact, in one of the trials, the participant toppled the obstacle with the passive leg (dotted curve). The maximum toe height reached with the passive leg was 0.106 m, whereas the powered leg reached 0.231 m.

\begin{figure}
	\centering
	\includegraphics[width=.8\linewidth,trim={1cm 0cm 2cm .5cm},clip]{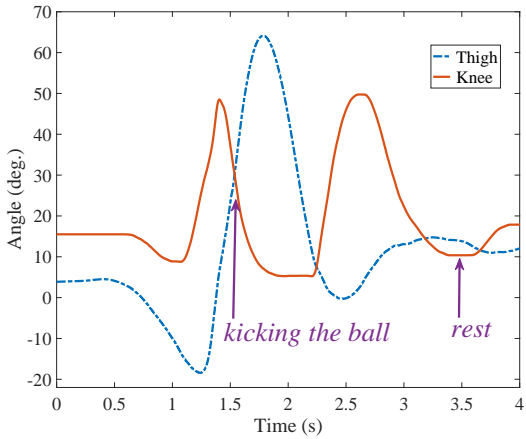}
	\caption{Thigh and knee angles during kicking a soccer ball with the powered prosthesis. After the shot, the leg retracts and then is put on the ground (rest).}
	\label{fig:SoccerBall}
\end{figure}

As the last non-rhythmic experiment, we tested the performance of kicking a soccer ball with the passive and powered legs. We chose ball velocity as our performance metric. The trials with the passive leg resulted in velocities of 1.4 and 1.9 m/s (mean: 1.65 m/s), whereas the powered leg produced velocities of 5.0 and 5.5 m/s (mean: 5.25 m/s). The use of the powered prosthesis allows the participant to kick the soccer ball with a substantially higher velocity (218\% increase in the mean ball velocity) when compared to using the passive prosthesis. Fig. \ref{fig:SoccerBall} shows the thigh and knee angles as the subject kicks the ball. Notice that the ball is kicked before maximum hip flexion (and hence maximum knee extension), but due to inertias, the leg continues moving forward. After reaching maximum flexion, the thigh retracts and the knee flexes for ground clearance. Finally, the thigh slightly extends forward, causing the knee to extend and the leg to rest on the ground. This shows the benefit of designing knee and ankle controllers based on following the motion of the thigh, which allows the subject to manage all of these maneuvers without difficulty.

\subsection{Rhythmic Tasks}
\begin{figure}
	\centering
	\subfigure[]{\includegraphics[width=.97\linewidth]{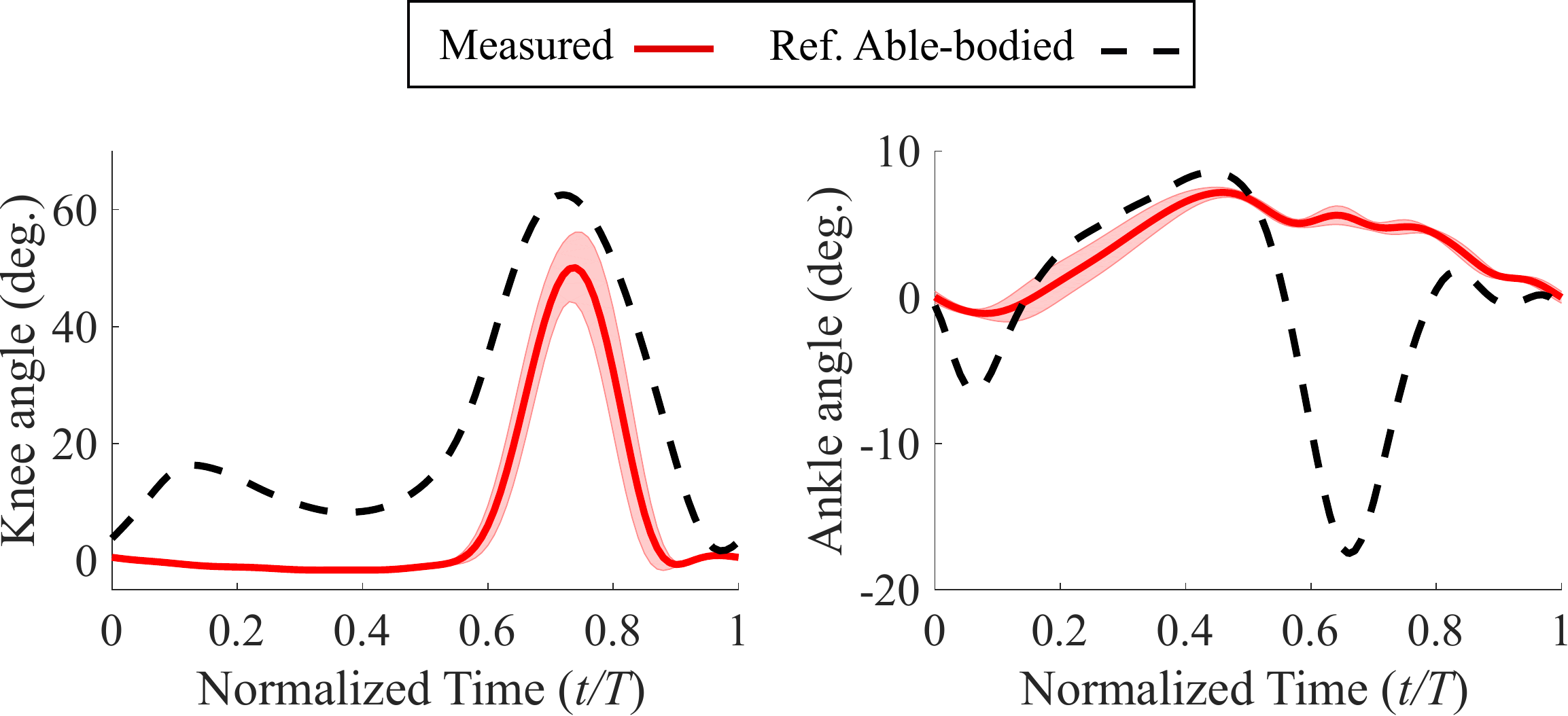}\label{fig:treadmill1x5passive}}
	\subfigure[]{\includegraphics[width=.97\linewidth]{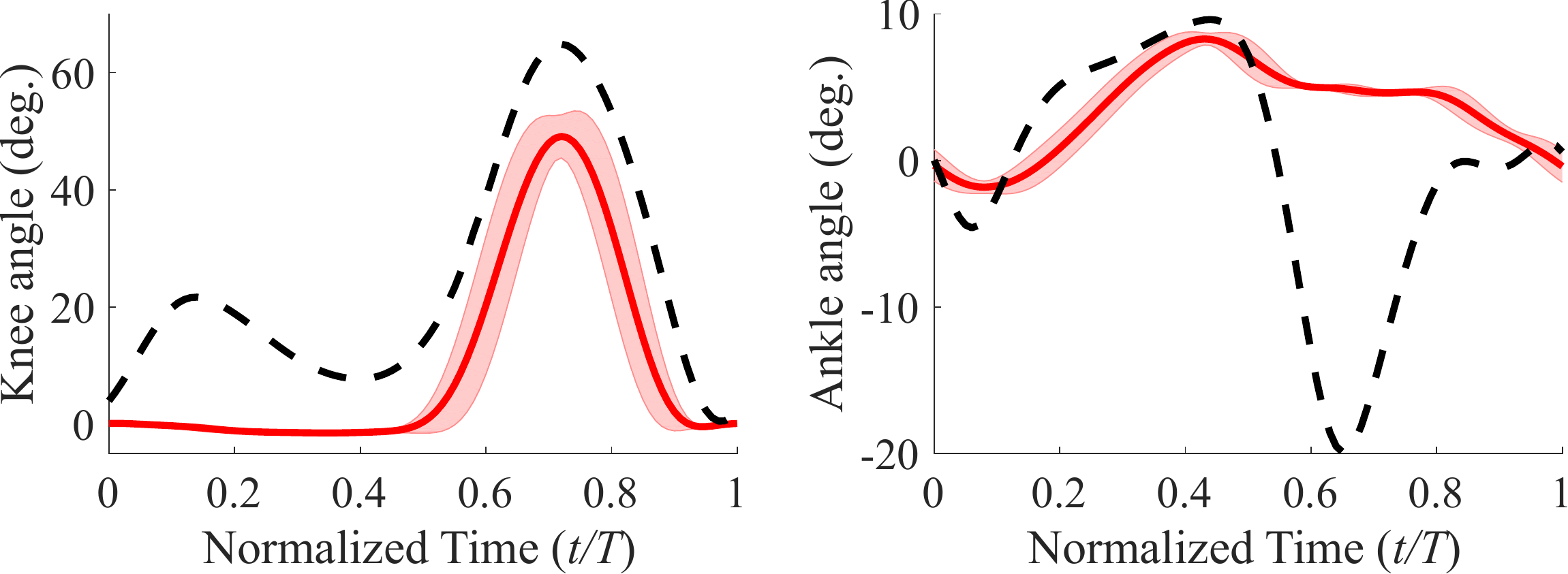}\label{fig:treadmill2x2passive}}
	\subfigure[]{\includegraphics[width=.97\linewidth]{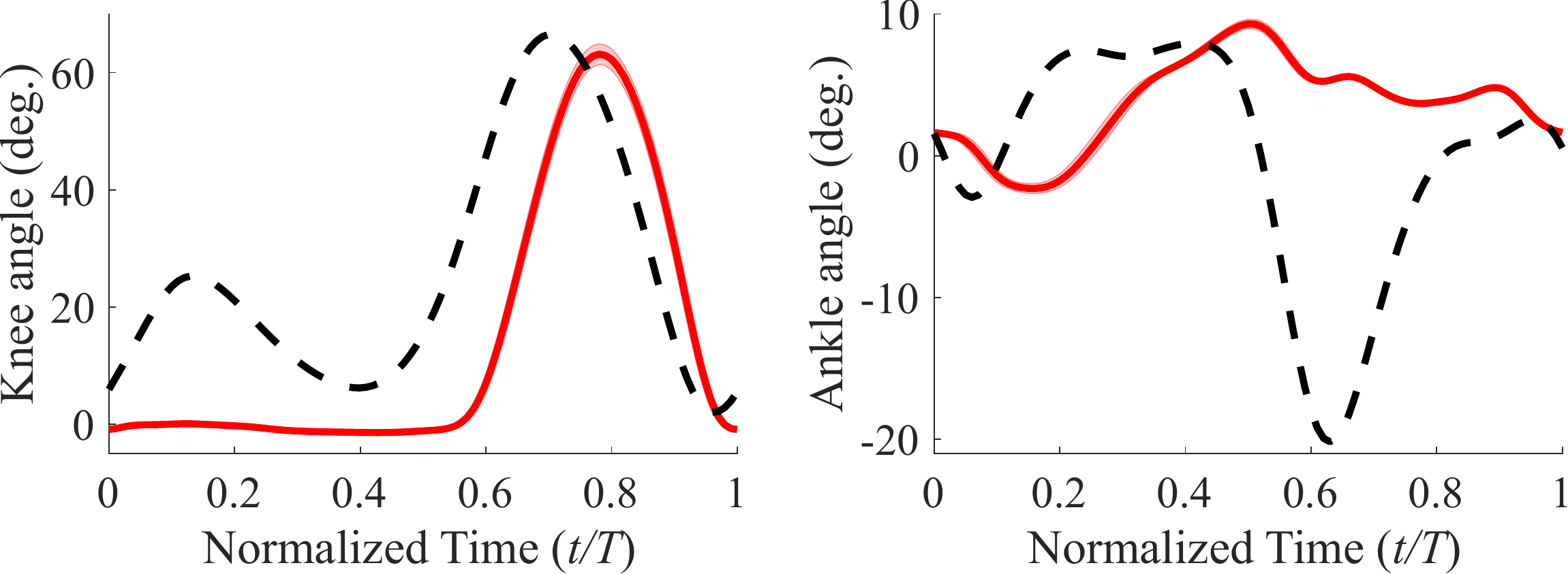}\label{fig:treadmill3x0passive}}
	\caption{Mean $\pm$ std of the joint angles of the passive leg as a function of normalized time for the treadmill tests with the amputee subject; \subref{fig:treadmill1x5passive} a 60-second trial with slow speed (0.7 m/s); \subref{fig:treadmill2x2passive} a 60-second trial with normal speed (1.0 m/s); and \subref{fig:treadmill3x0passive} a 45-second trial with fast speed (1.3 m/s). The able-bodied reference data for the three speeds are from \cite{Winter2005}.}
	\label{fig:treadmillPoltsPassive}
\end{figure}

\begin{figure}
	\centering
	\subfigure[]{\includegraphics[width=.97\linewidth,trim={1.5cm .7cm 1.5cm 0cm},clip]{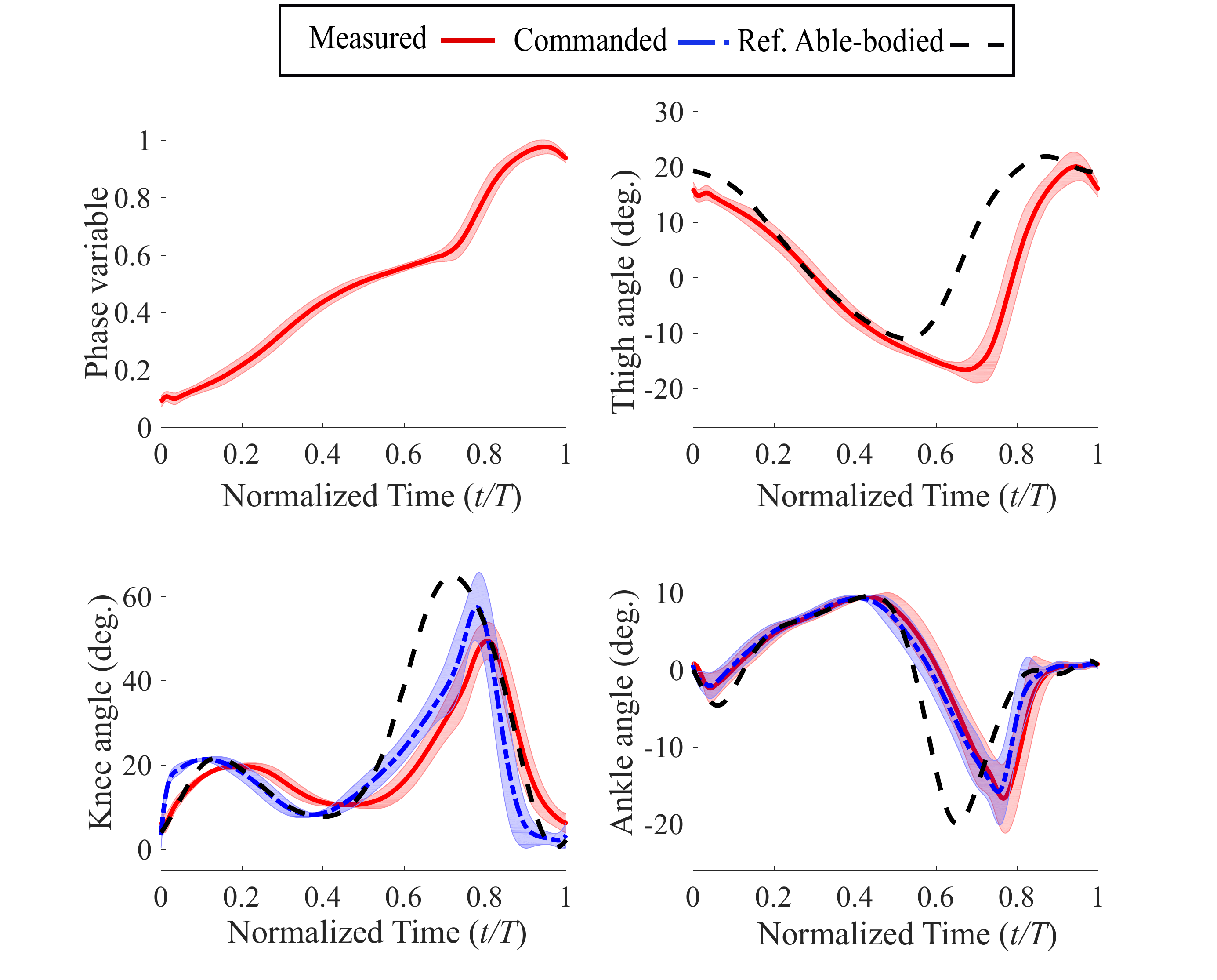}\label{fig:treadmill1x5}}
	\subfigure[]{\includegraphics[width=.97\linewidth,trim={1.5cm 1.2cm 1.5cm .5cm},clip]{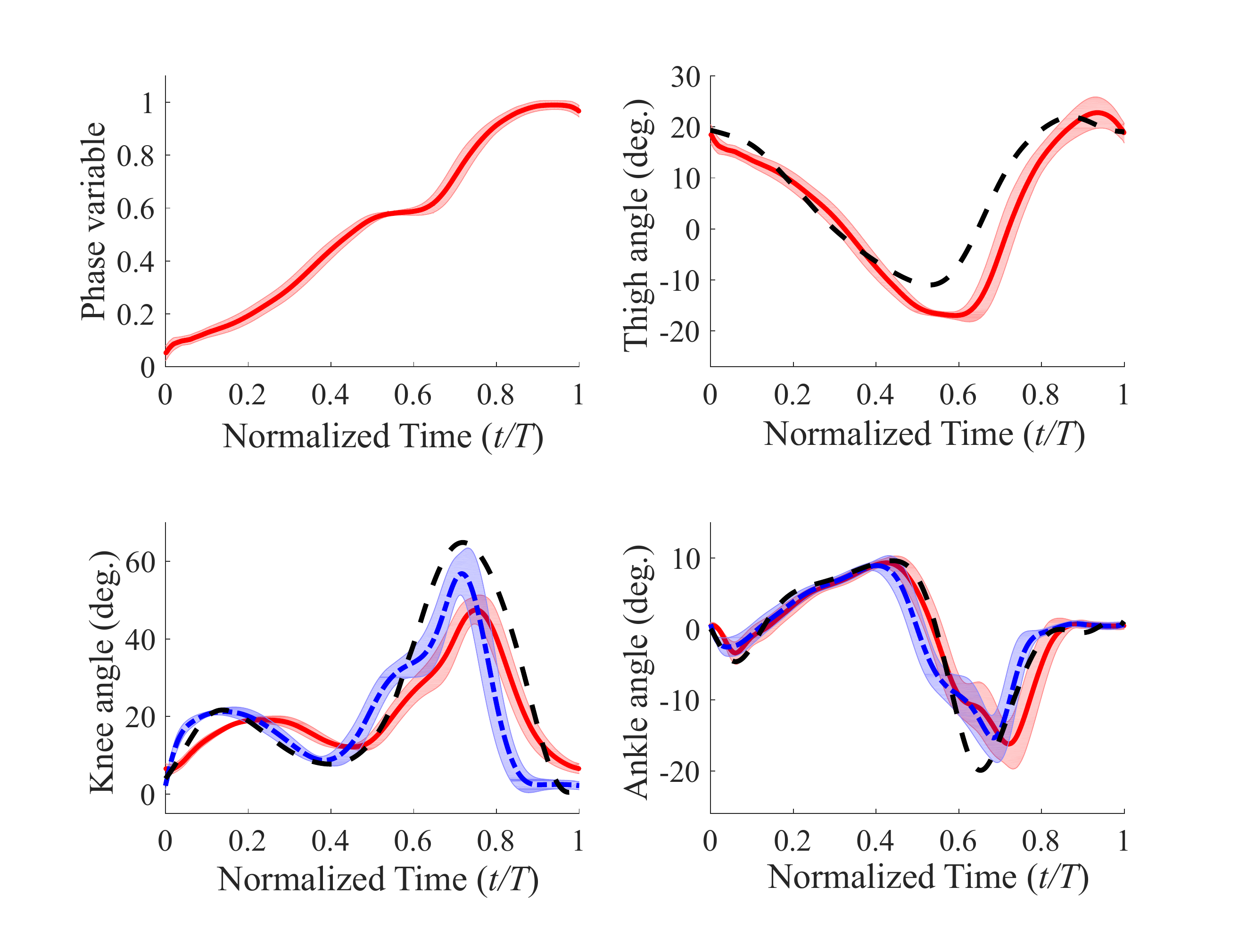}\label{fig:treadmill2x2}}
	\subfigure[]{\includegraphics[width=.97\linewidth,trim={1.5cm 1.4cm 1.5cm .5cm},clip]{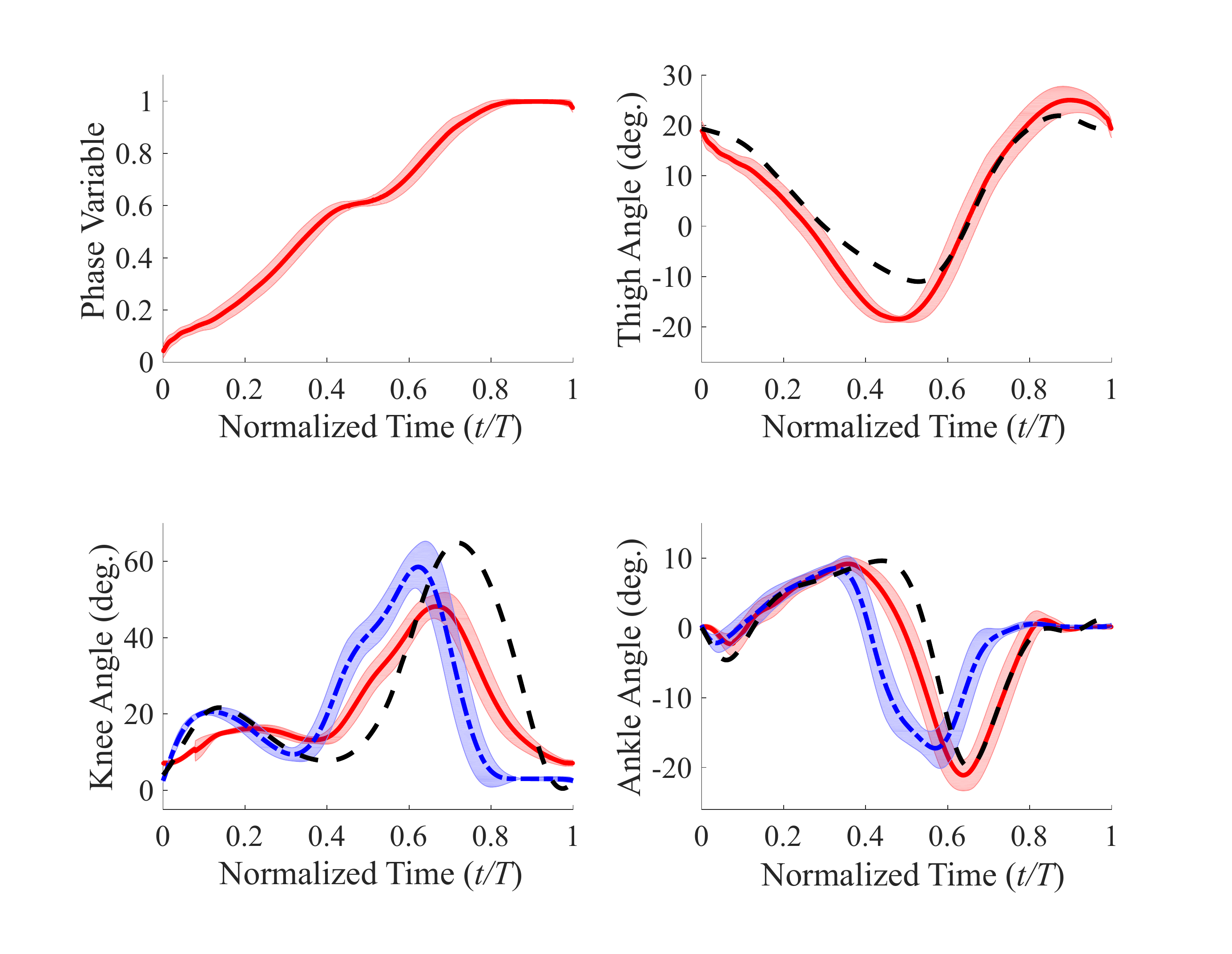}\label{fig:treadmill3x5}}
	\caption{Mean $\pm$ std for phase variable, and commanded and measured joint angles vs. normalized time for the treadmill tests with the amputee subject; \subref{fig:treadmill1x5} a 60-second trial with slow speed (0.7 m/s); \subref{fig:treadmill2x2} a 60-second trial with normal speed (1.0 m/s); and \subref{fig:treadmill3x5} a 30-second trial with maximum speed (1.6 m/s). Reference trajectories are normal-speed in all cases.}
	\label{fig:treadmillPolts}
\end{figure}

\begin{figure}
	\centering
	\subfigure[]{\includegraphics[width=0.96\linewidth,clip]{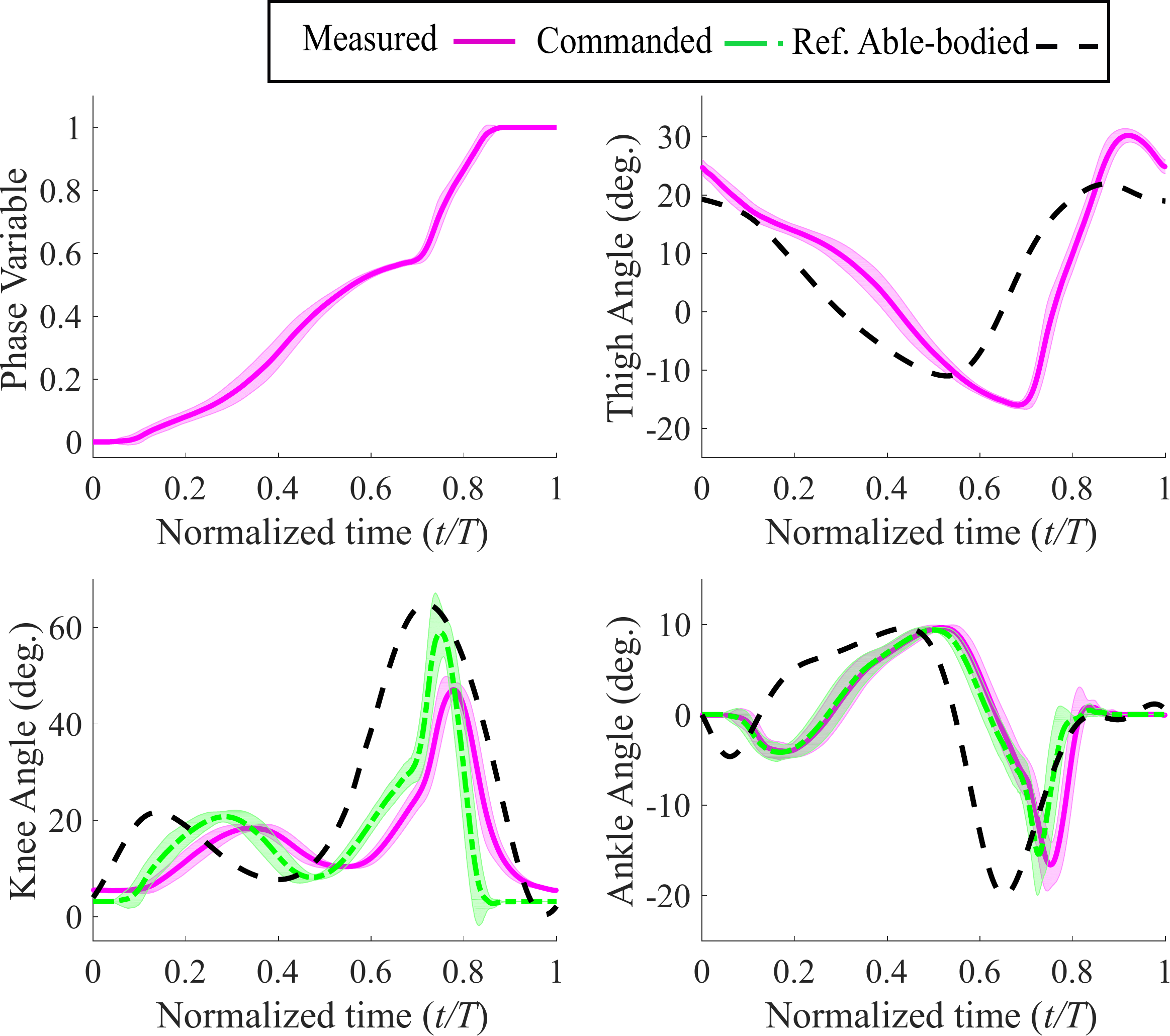}\label{fig:treadmill1x5able}}
	\subfigure[]{\includegraphics[width=0.96\linewidth,clip]{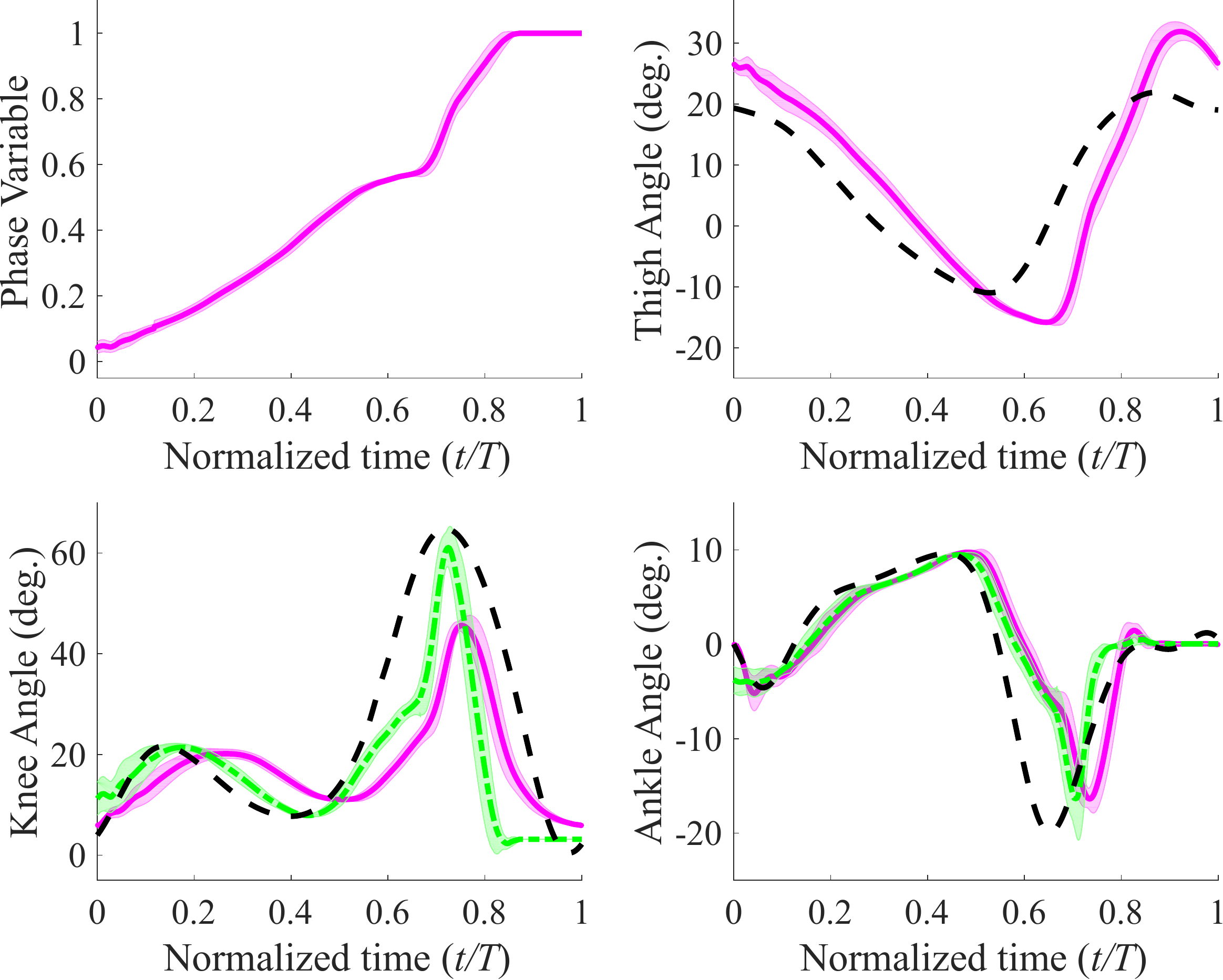}\label{fig:treadmill2x0able}}
	\subfigure[]{\includegraphics[width=0.96\linewidth,clip]{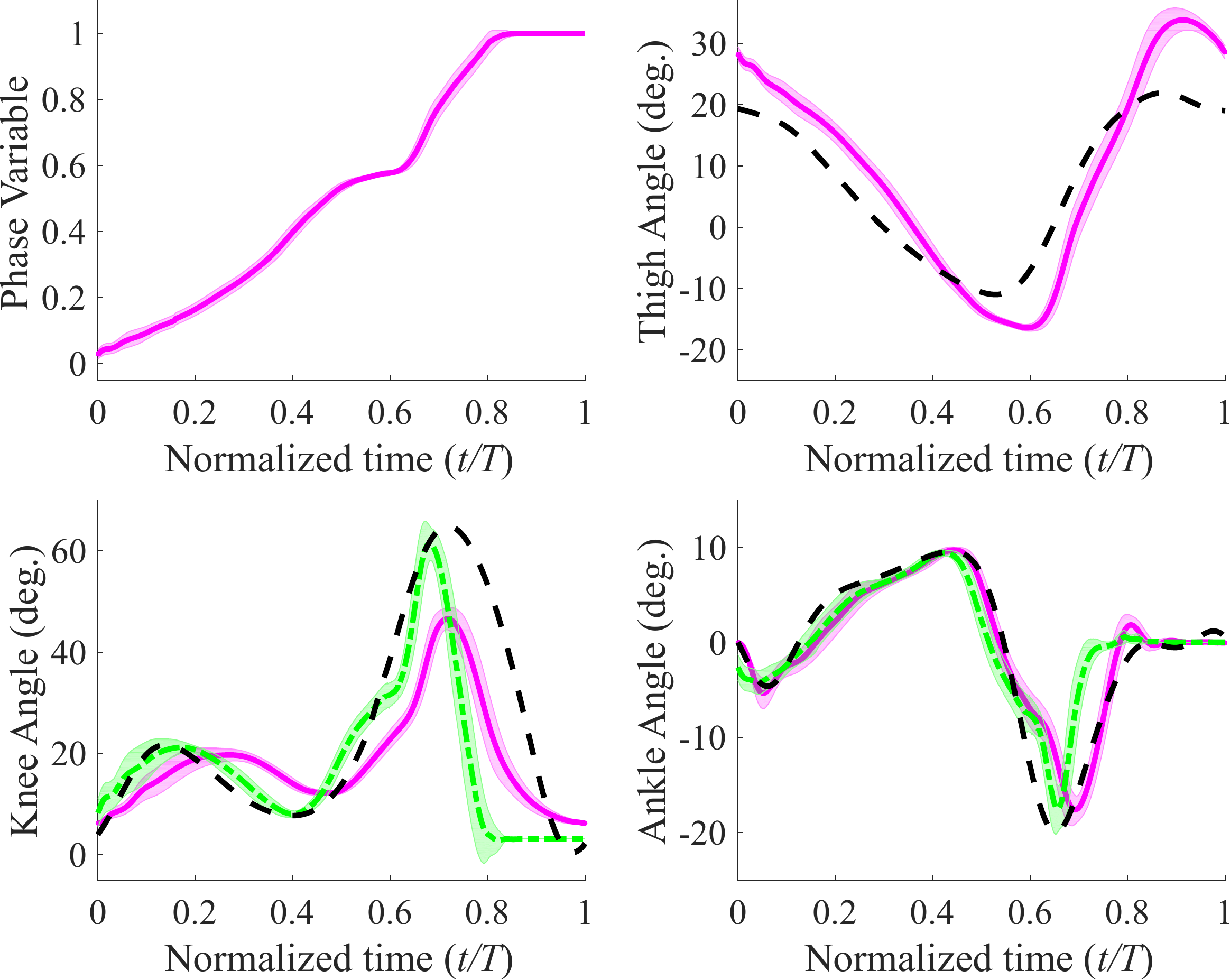}\label{fig:treadmill2x5able}}
	\caption{Mean $\pm$ std for phase variable, and commanded and measured joint angles vs. normalized time for treadmill test with the able-bodied subject; \subref{fig:treadmill1x5able} a 30-second trial with slow speed (0.7 m/s); \subref{fig:treadmill2x0able} a 30-second trial with normal speed (0.9 m/s); and \subref{fig:treadmill2x5able} a 30-second trial with fast speed (1.1 m/s). Reference trajectories are normal-speed in all cases.}
	\label{fig:treadmillPoltsAble}
\end{figure}

The knee and ankle angle trajectories for the trials with the passive leg on the treadmill are shown in Fig. \ref{fig:treadmillPoltsPassive}. In particular, the absence of ankle plantarflexion (pushoff) in these trials is noticeable. In contrast, the ankle pushoff is quite conspicuous in the powered leg results of Figs. \ref{fig:treadmill1x5} to \ref{fig:treadmill3x5}. Note that since $q_h^0$ and $q_h^{min}$ were not changed in these trials, the minimum thigh angle is reached later than the reference trajectory during slow walking (Fig. \ref{fig:treadmill1x5}). In other words, the ratio of stance to swing duration increases in order to provide extra time to achieve the minimum thigh angle while the foot is constrained to follow the treadmill speed. As the treadmill speed increases, the minimum thigh angle shifts to the left (Figs. \ref{fig:treadmill2x2} and \ref{fig:treadmill3x5}) and the stance to swing duration ratio decreases. Also, the amplitude of the minimum thigh angle is consistently larger than the reference (about $-17^{\circ}$ for all three speeds as opposed to $-11^{\circ}$ for the reference trajectory). This means that the ankle pushoff was not fast enough to quickly reverse the direction of motion of the thigh and prepare it for the swing phase \cite{Lipfert2014}. Similar trends are visible in the results of experiments with an able-bodied subject with the same control parameters (Figs. \ref{fig:treadmill1x5able} to \ref{fig:treadmill2x5able}). The Pearson correlation coefficients for the knee joint (Table \ref{tab:correlation}) confirm these observations. The overlong stance to swing duration ratio in slow speed with the powered leg shows its effect in the significant lag in both knee and ankle trajectories compared to the normative gait (Fig. \ref{fig:treadmill1x5}). This lag increases the difference between the normative and measured knee trajectories, which in turn decreases the correlation compared to that of the passive leg (0.80 versus 0.93). However, with the increase of speed, the correlation of the powered leg improves, and in fast speed it significantly outperforms the passive leg (0.95 versus 0.75). These observations will be discussed in detail in the next section.

\begin{table}
	\caption{Pearson correlation coefficients between the joint trajectories of each prosthetic leg (passive or powered) and the speed-matched normative able-bodied data for different walking speeds \cite{Winter2005}.}
	\label{tab:correlation}
	\begin{center}
		\renewcommand{\arraystretch}{1.5} 
		\begin{tabular}{|c||c|c|c|c|} \hline
			
			& \textbf{Pas knee}   & \textbf{Pwr knee}   &\textbf{Pas ankle}  &\textbf{Pwr ankle}\\ \hline
			
			\textbf{Slow}                & 0.93    & 0.80   & 0.15  & 0.71  \\ \hline
			\textbf{Normal}                & 0.96    & 0.96  & 0.16  & 0.94 \\ \hline
			\textbf{Fast}                & 0.75   & 0.95  & -0.20  & 0.97 \\ \hline
			
		\end{tabular}
		\renewcommand{\arraystretch}{1}
	\end{center}
\end{table}

\begin{figure}
	\centering
	\subfigure[]{\includegraphics[width=.95\linewidth]{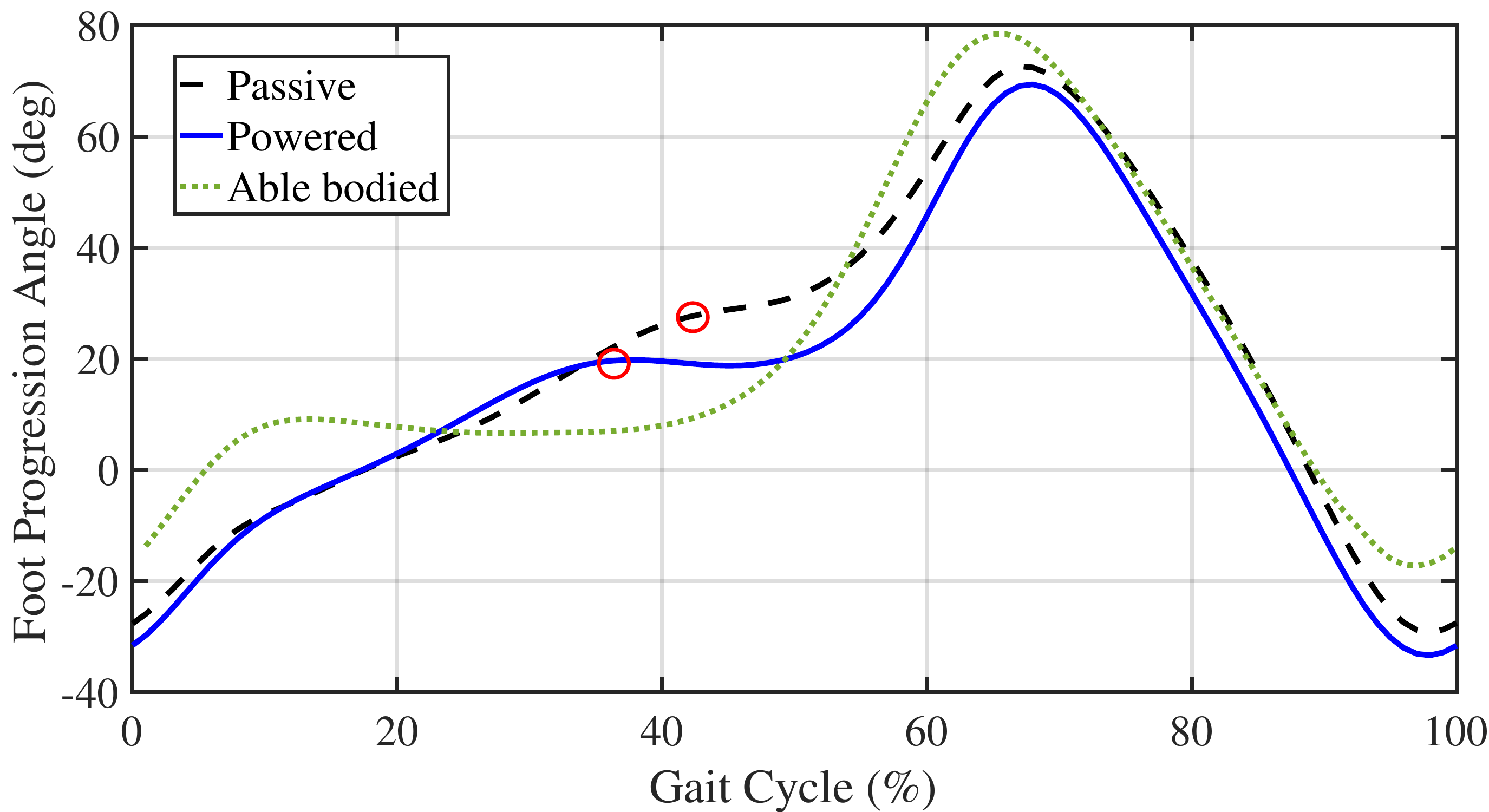}\label{fig:vaulting_compared}}
	\subfigure[]{\includegraphics[width=1\linewidth]{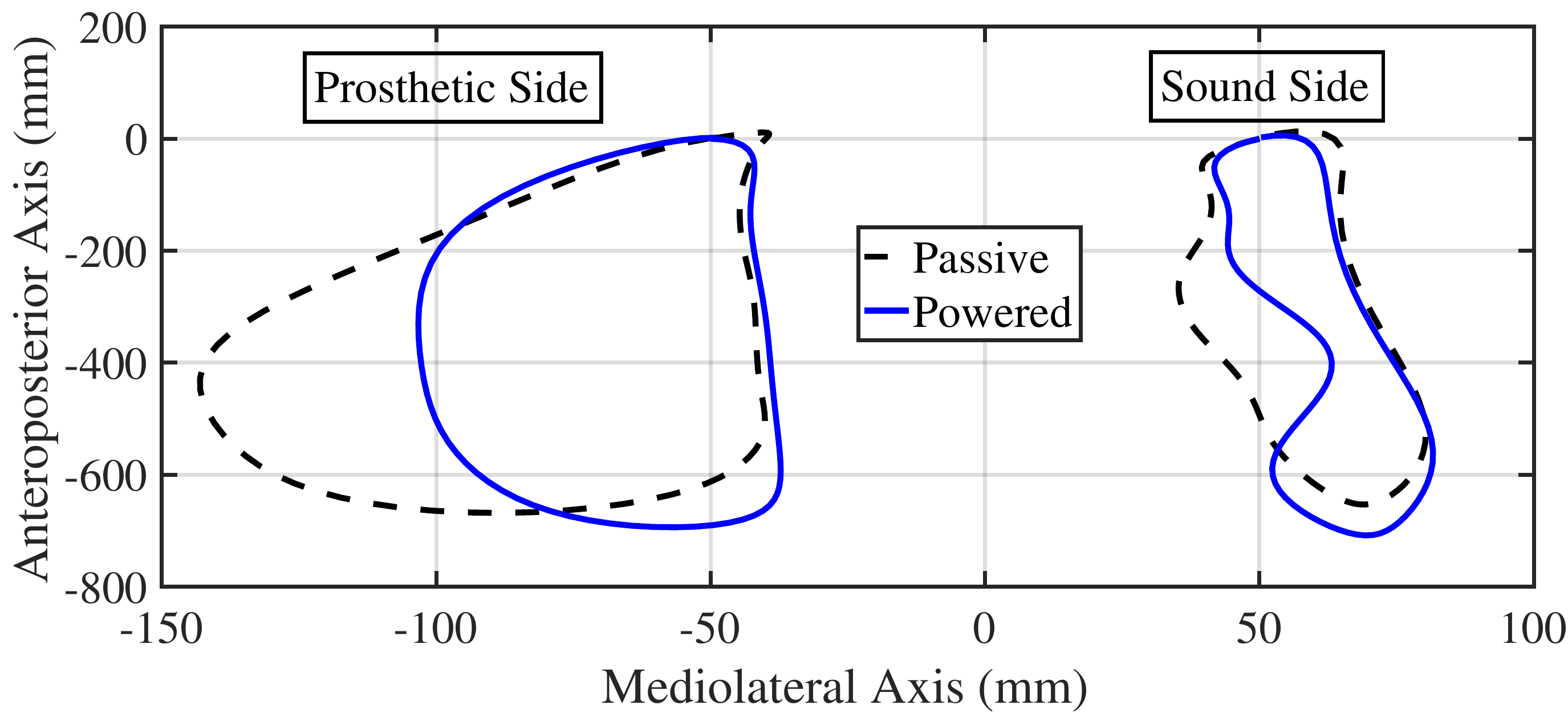}\label{fig:circumduction}}
	\caption{\subref{fig:vaulting_compared} Mean foot angles of the amputee subject's sound leg while wearing the powered (blue line) and passive (black dashed line) legs during fast treadmill walking. Also shown is the mean global foot angle of able-bodied reference data at fast cadence (green dotted line) \cite{Winter2005}. Red circles show peak global foot angle during single support period, i.e., vaulting angle. \subref{fig:circumduction} Mean ankle marker trajectories during fast treadmill walking representing differences in hip circumduction while using the powered and passive legs. Left: prosthetic leg side. Right: sound leg side.}
	\label{fig:compensations}
\end{figure}

Next, we focus our analysis on common foot clearance compensation strategies employed by transfemoral amputees (i.e., vaulting and hip circumduction) for walking tasks. The goal is to determine whether our controller improves these compensations compared to the subject's passive leg. Fig. \ref{fig:vaulting_compared} shows the mean global foot angles of the sound leg during the fast treadmill walking speed using passive and powered legs. Additionally, the global foot angle of the able-bodied reference data is shown for comparison \cite{Winter2005}. It can be seen that the amputee subject employed a rising global foot angle during the single-support phase (for both passive and powered legs) when compared to the much more constant angle of able-bodied reference data during the same phase. Importantly, the vaulting angle (the angle at which the velocity crosses zero, indicated by a red circle) was lower when using the powered leg when compared to using the passive leg. See Table \ref{tab:vaulting} for mean and SD values. Note that the markers on the sound leg were not moved during the entire experiment (with passive and powered legs), and thus there was not any possibility of marker placement error. 

\begin{table}
	\caption{Mean (SD) values of vaulting (deg.) for powered and passive legs at slow, normal and fast walking speeds using the global foot angle method.}
	\label{tab:vaulting}
	\begin{center}
		\renewcommand{\arraystretch}{1.5} 
		\begin{tabular}{|c||c|c|c|c|} \hline
			
			& \textbf{Passive}   & \textbf{Powered}   \\ \hline
			
			\textbf{Slow}                & 16.1 (1.3)    & 13.6 (1.6)   \\ \hline
			\textbf{Normal}                & 21.0 (1.8)    & 13.6 (1.4)  \\ \hline
			\textbf{Fast}                & 28.6 (1.7)   & 20.3 (2.0)   \\ \hline

		\end{tabular}
		\renewcommand{\arraystretch}{1}
	\end{center}
\end{table}

Next, we present the results of hip circumduction. Fig. \ref{fig:circumduction} shows mean ankle marker trajectories during fast treadmill walking. A clear reduction in the medio-lateral range of motion of the prosthetic ankle marker trajectories can be seen when using the powered leg compared to passive. Furthermore, circumduction symmetry ratios were improved with the powered leg for all speeds, showing that the amount of circumduction with the powered leg was closer to that of the sound leg (taken as a reference). See Table \ref{tab:circumduction} for mean, SD, and symmetry index values. 

\begin{table}
	\caption{Ankle lateral deviation (hip circumduction): powered versus passive prostheses mean (SD) at different speeds.}
	\label{tab:circumduction}
	\begin{center}
		\renewcommand{\arraystretch}{1.5} 
		\begin{tabular}{|c||c|c|c|c|} \hline
			
			& \textbf{Passive}   & \textbf{Powered}   &\textbf{SI Passive}  &\textbf{SI Powered}\\ \hline
			
			\textbf{Slow}                & 46.4 (10.2)    & 45.5 (5.7)   & 0.2  & 0.1  \\ \hline
			\textbf{Normal}                & 71.8 (19.4)    & 55.8 (13.5)  & 0.3  & 0.1 \\ \hline
			\textbf{Fast}                & 105.8 (9.5)   & 70.3 (12.1)  & 0.8  & 0.5 \\ \hline

		\end{tabular}
		\renewcommand{\arraystretch}{1}
	\end{center}
\end{table}


\section{Discussion}
\label{sec:Discussion}
As the supplemental video presents, the proposed controller enabled the amputee subject to accomplish a number of non-rhythmic (stepping forward and backward, instantaneous start and stops, walking over obstacles, kicking a soccer ball) and rhythmic (walking on a treadmill with different speeds) tasks. Although the designed virtual constraints were based on a rhythmic task, i.e., normal-speed walking kinematics, the holonomic nature of the controller helped the subject perform non-rhythmic tasks as well. Unlike previous controllers that used thigh angle to parameterize the gait \cite{Villarreal2017b,Quintero2018}, the proposed controller is not limited to only one type of motion (rhythmic or non-rhythmic). In what follows, we discuss about the performance of the proposed controller and the observed improvements compared to that of the amputee subject's passive leg. Since the goal of this study has been to investigate the acute biomechanical effects of the powered leg, we excluded the variables that are significantly dependent on adaptation time, such as step width \cite{Schmalz2014}.

\subsection{Biomechanical Analysis of Rhythmic and Non-Rhythmic Tasks}

One of the main objectives of this work is to demonstrate the ability of the powered prosthesis to facilitate both rhythmic and non-rhythmic tasks. Compensatory mechanisms during walking (primarily for aiding foot clearance) are problematic in general because of higher energy expenditure \cite{schmalz2002energy,waters1999energy,atlas2004energy} and deterioration of the intact joints \cite{gailey2008review,kulkarni1998association,struyf2009prevalence}. We hypothesized that the assisted (powered) ankle plantarflexion during push-off and knee flexion during early swing would help reduce these compensations, namely, vaulting and hip circumduction.

Our amputee participant exhibited moderate to severe vaulting with his passive leg. Analysis with the vaulting angle metric shows that vaulting improved with the powered leg as compared to passive for all walking speeds tested. Based on this result, we expect that the use of our powered device with the proposed controller can mitigate the deleterious effects of vaulting such as metatarsal pain and the sinus tarsi syndrome caused by overloading of the forefoot \cite{Drevelle2014}. Moreover, we can expect a reduction in energy expenditure through reduced plantarflexion and vertical displacement of the whole body \cite{waters1999energy}. Likewise, the use of the powered leg led to a clinically significant reduction in hip circumduction compared to the excessive level observed when using the passive leg. Similar benefits to those of vaulting reduction can be expected from the hip circumduction improvement as well.

The joint trajectories of the powered prosthesis had noticeably greater correlations with the reference able-bodied data compared to those of the passive leg. The passive leg only had a greater correlation at the knee for the slow speed, which was due to the powered leg's use of a single parameterization for all speeds, as discussed in the previous section. As the speed increased, the greater propulsion force generated from the ankle pushoff of the powered leg resulted in more normative trajectories than the passive leg.

The phase variables reported in Figs. \ref{fig:handrailFWD} (overground forward walking), \ref{fig:handrailBWD} (overground backward walking), \ref{fig:treadmillPolts}, and \ref{fig:treadmillPoltsAble} (treadmill walking) saturate at the end of the swing phase. This means that the touchdown thigh angles for both subjects have been greater than the default value $q_h^0$. As mentioned in Section \ref{sec:ControlDesign}, this saturation can be eliminated through tuning of $q_h^0$. However, even without tuning (as was the case in the conducted experiments), the saturation did not seem to have a noticeable effect on the performance. Since the touchdown thigh angle is controlled by the subjects, the chosen angle can be considered as the angle that is preferred by them to attain the best performance and comfort.

The commanded swing knee extensions in both Figs. \ref{fig:treadmillPolts} and \ref{fig:treadmillPoltsAble} are faster than the normative trajectory. This happens because in the designed virtual constraint, $s=1$ corresponds to the maximum thigh angle during swing, which occurs at 87\% of the gait cycle and not at 100\% (see Fig. \ref{fig:humanAngles}(a)). This leads to a shorter swing phase, which in turn results in a faster knee extension commanded trajectory compared to the reference trajectory (Figs. \ref{fig:treadmillPolts} and \ref{fig:treadmillPoltsAble}). In practice, due to the natural dynamics of the actuator and the leg, and the lag resulting from these factors, the measured knee extension was more similar to the reference knee angles (Figs. \ref{fig:treadmillPolts} and \ref{fig:treadmillPoltsAble}). In any case, this difference can be eliminated through changing the tunable parameter $c$ in \eqref{eq:stancePV}.

The joint ankle power of the powered leg (as depicted in Fig. \ref{fig:anklePower}) is in the same level as reported in \cite{Quintero2018} (about 100 W), and as expected, substantially increases pushoff power compared to the passive leg. In the case of passive prostheses, the lack of active pushoff and knee flexion during the stance phase results in reduced propulsion. Therefore, to maintain a constant walking velocity, braking forces are usually reduced in the subsequent stance phase \cite{adamczyk2015mechanisms}. This asymmetry contributes to numerous clinical symptoms, for example, osteoarthritis and back pain \cite{Devan2014}. Thus the improvement in symmetric use of the legs through increase of the ankle pushoff power can lead to reducing common overuse symptoms. The peak joint power is mainly limited by the motor's speed and torque range as well as the high reflected inertia of the actuator system, which consumes a substantial part of the motor power for acceleration and deceleration. These factors also limit the knee flexion speed at the swing phase onset, which resulted in toe-stubbing at some strides. We expect that with the use of higher-power motors with lower reflected inertias (i.e., lower transmission ratios) such as \cite{Elery2018}, the controller will achieve pushoff powers comparable to that of able-bodied subjects.

The speed-invariance of the controller and its improved pushoff management also allowed for greater walking speeds compared to the maximum speed achieved in \cite{Quintero2018}, which was 1.2 m/s. With the present controller, the amputee subject was able to comfortably walk at 1.6 m/s, which can also be compared with the maximum walking speeds achieved in \cite{Lenzi2014} (1.4 m/s) and \cite{Jayaraman2018} (1.6 m/s). It is worth noting that the subject who achieved 1.6 m/s in \cite{Jayaraman2018} was able to walk at the same speed using his passive leg, whereas our amputee subject was only able to reach a maximum speed of 1.3 m/s with his passive leg. In our preliminary experiments, the able-bodied subject wearing the prosthesis using a bypass was able to reach 1.8 m/s with the present controller. The limitation for high speeds primarily originates from the torque saturation of the motors and not from the controller. The inadequate ankle torque and power during stance and the limited speed of the knee actuator during swing require the users to compensate using torque from their hips, which quickly leads to fatigue at fast speeds. Overall, the results highlight the substantial improvement made in the controller performance as compared to \cite{Villarreal2017b} in which pushoff had to be eliminated to avoid abrupt jumps.

Backward walking is an interesting feature of our control approach that cannot be achieved with the traditional impedance-based FSM control methods of powered knee-ankle prostheses, which are limited to working only in the forward direction. As mentioned, for the backwards walking task, our participant utilized a step-to gait when using the passive prosthesis. The lead leg was the sound leg, which the prosthetic leg followed. This choice of gait is not surprising as the passive knee is incapable of load bearing when behind the body. In contrast, a more natural step-through gait was adopted with the powered prosthesis. Comparison of step symmetry for backwards walking shows that the powered leg enables more stable and more natural-looking maneuvers. The phase variable in backward walking is not monotonic at all points (see Fig. \ref{fig:handrailBWD}). The break in monotonicity around $s = 0.5$ corresponds to the minimum thigh angle, i.e., when the thigh angle is in its most backward position, preparing for touchdown. As can be seen from the video, the amputee subject hesitates when putting his leg on the ground in order to ensure that the knee is fully extended and the prosthesis is ready for weight bearing. This can be attributed to the short acclimation time with the powered leg and especially this particular task, which cannot be accomplished with passive legs.

An analysis of the obstacle crossing task (Fig. \ref{fig:obstacle}) shows that our participant was easily able to clear the obstacle using his powered leg, with a clearance of 100 mm between the toe marker and top edge of the obstacle. While using the passive leg, this same task was extremely difficult. The obstacle was toppled over on one attempt and was barely cleared on the second try. The controller of the powered leg is designed to provide maximum knee flexion (clearance) at a hip angle of approximately $12^{\circ}$ of flexion. Therefore, the optimal way to cross the obstacle for an amputee would be to maintain this hip angle and guide the hip joint (and thus the prosthesis) over the obstacle. Our amputee participant was quickly able to figure out this optimal strategy without any guidance from the experimenters. This is due to the holonomic mapping between the thigh angle and the foot position, which is easy for the subjects to learn. Once again, this emphasizes the intuitiveness of the controller for common volitional tasks. 

Kicking the soccer ball requires a rapid yet controlled extension of the knee. Our participant was able to kick the soccer ball using the passive as well as the powered legs. However, the mean ball velocity was more than three times higher when using the powered leg. This result is impressive as it demonstrates the intuitive control of the powered knee and ankle at high speeds. Based on the control strategy, a fast extension of the knee can be achieved by a rapid flexion of the hip. Our amputee participant was able to gain an understanding of these mechanics and execute them after only a couple of practice trials. Again, no guidance or feedback was provided to the participant regarding the optimal mechanics.

\subsection{Limitations}
The use of a purely holonomic set of virtual constraints also has a limitation. There is a relatively flat section in the middle part of the phase variable plots (normalized time of about 0.5-0.6) in both Figs. \ref{fig:treadmill1x5} to \ref{fig:treadmill3x5} and Figs. \ref{fig:treadmill1x5able} to \ref{fig:treadmill2x5able}, meaning that the rate of the phase variable is almost zero. To investigate the reason for this phenomenon, note that
\begin{equation}
\label{Eq:singularity}
\dot{s}=\frac{ds}{dq_h}\dot{q}_h,
\end{equation}
\noindent which means for $\dot{q}_h=0$, we will have $\dot{s}=0$. This condition occurs during pushoff (transition from S2 to S3). As a result of $\dot{s}=0$, the knee and ankle rates also tend to vanish, and pushoff becomes slower. This contributes to the thigh continuing backward, before the ankle plantarflexion increases enough to stop the backward motion and drives the thigh forward. This is intrinsic to the holonomic virtual constraints and can be regarded as a trade-off.

The proposed controller has a few more tunable parameters than the previous version of the phase-based controller \cite{Quintero2018}. However, as we showed, both the able-bodied and the amputee subjects were able to comfortably walk with the default parameters (taken from the normative human kinematics data). The addition of tunable parameters provides the ability of easily adopting various walking styles, such as different step lengths or pushoff initiation points, which was not possible in \cite{Quintero2018}. See \cite{Quintero2018Interface} for an example of using these tunable parameters in an intuitive clinical control interface.

The amputee subject tended to use the handrails more in the experiments with the powered leg. This can be mostly attributed to the acclimation factor. This hypothesis is supported by the fact that the subject did use the treadmill handrails in slow and fast speeds with his passive leg and only in his self-selected normal speed he did not rely on them. In any case, we do not expect that the use of handrails would affect our results, because 1) at slow and fast speeds the subject used the handrails both with passive and powered legs, and 2) the results at normal speed (at which the handrails were used with the powered leg, but not with the passive leg) show no significant deviation in the trend from slow to fast speed.

Another observation from the treadmill tests was the stance to swing duration ratio. Note that the leg's joint kinematics and especially the maximum and minimum of thigh angle change as walking speed varies \cite{Kirtley1985}. For the present study, we kept the kinematics (virtual constraints) unchanged, in order to demonstrate the ability of the controller to work in different situations with minimal tuning. As a result, for low speeds the stance to swing duration ratio was greater than expected. However, the decrease of stance to swing ratio and earlier pushoff in higher speeds is still in accordance with observations from able-bodied subjects \cite{Kirtley1985}. Moreover, walking speed can be estimated using fairly straightforward methods (see \cite{Lenzi2014,Quintero2017}, for example), and the virtual constraints can be changed accordingly. A similar idea can incorporate the necessary kinematic differences between forward and backward trajectories. Note that adding these additional virtual constraints is merely a  kinematic modification, whereas the dynamic joint attributes (i.e., joint impedances) remain unchanged. This provided invariance across subjects for the phase-based controller that was proposed and tested in our previous work \cite{Quintero2018}. The comparison between able-bodied and amputee users of the powered prosthesis (Figs. \ref{fig:treadmillPolts} and \ref{fig:treadmillPoltsAble}) offer some evidence that our updated phase variable retains this beneficial property. However, the purpose of this study was to demonstrate the potential for increased functional abilities rather than subject independence.


\section{Conclusion}
A new control framework for a range of rhythmic and non-rhythmic tasks was designed for powered knee-ankle prostheses and validated through experiments with an above-knee amputee subject. The controller uses a phase variable defined as a piecewise holonomic function of the thigh angle with transitions based on a finite state machine. Unlike the previous phase-variable-based controllers (see \cite{Quintero2018}), the specific design of the presented phase variable enables the subjects to manage both rhythmic and non-rhythmic tasks, which makes it more practical for real-world applications. Through the experiments with an amputee subject, it was shown that the controller enabled operation in an extensive range of walking speeds with clinically significant improvements in compensations associated with passive prostheses.

Although the controller facilitates a wider range of tasks than walking, it does not encapsulate tasks such as stair ascent and descent. However, the structure provided is flexible for embodying new sets of kinematics in task-recognition frameworks \cite{Varol2010,Bartlett2017}. Similarly, a paradigm that unifies different speeds and inclines (as in \cite{Embry2018}) can be incorporated to expand the range of tasks that the proposed controller can manage, especially for walking on non-flat surfaces.

Another interesting extension of the controller would include a nonholonomic correction for the interruption in pushoff for a faster and smoother transition to the swing phase. Furthermore, we plan to test the controller on our newly designed leg \cite{Elery2018}, which provides greater torques as well as backdrivability. These investigations will provide a better understanding of the controller's additional capabilities and its associated benefits.

\section*{Acknowledgment}
The authors would like to thank Christopher Nesler for his help in conducting the experiments.


\bibliography{Bibliography}
\bibliographystyle{IEEEtran}


\end{document}